\documentclass[preprint,3p,times]{elsarticle}

\usepackage{amssymb}

\usepackage{booktabs}
\usepackage{multirow}
\usepackage{subfig}
\usepackage{amsmath}
\usepackage[dvipsnames]{xcolor}

\usepackage{enumitem}
\usepackage{url}

\journal{Pattern Recognition}

\begin{document}

\begin{frontmatter}



\title{Unraveling Machine Behavior \\by Multi-Level Bias Analysis and Detection: \\ Methodology and Application to Computer Vision}


\author[label1]{Ignacio Serna\corref{cor1}}
\author[label2,label3]{Aythami Morales}
\author[label2]{Julian Fierrez}
\affiliation[label1]{organization={Center for Humans and Machines, Max Planck Institute for Human Development},
            city={Berlin},
            country={Germany}}
\affiliation[label2]{organization={BiometricsAI, UAM},
            city={Madrid},
            country={Spain}}
\affiliation[label3]{organization={Department of Mathematics, ULPGC},
            country={Spain}}
\cortext[cor1]{Corresponding author: serna@mpib-berlin.mpg.de}

\begin{abstract}

Bias in machine learning, particularly in computer vision applications such as biometrics, is a critical issue with profound ethical, legal, and practical implications. This study investigates the presence and propagation of bias within Neural Networks through a comprehensive multi-level analysis spanning the learned latent space, layer activations, and the network's parameters. Based on this taxonomy, we propose three bias detection approaches: 1) SpaceBias (new method), which characterizes the latent space prior to the final classification layer using neighbor-probability distributions and quantifies bias with the two-sample Kolmogorov–Smirnov test on the per-group distributions. 2) ActivationBias (extension of the existing method InsideBias), which analyzes the activations of neural network filters and quantifies bias via a Mann–Whitney U test, based on the observed fact that underrepresented groups exhibit lower activation levels in the final convolutional layers. 3) WeightBias (extension of the existing method IFBiD), which uses a secondary neural network trained to identify biased patterns directly in the parameters of task-specific models. Unlike conventional methods, which assess neural network outcomes and treat the model as a black box, our proposed techniques provide insight into how biases manifest within the network architecture itself at different levels, offering a more nuanced and detailed understanding. Experiments are conducted on two complementary applications: gender classification in the DiveFace dataset (72,000 face images) and digit classification on a colored-MNIST benchmark with controlled bias severity. In total, more than 127,000 models with varying degrees and types of bias were trained and evaluated. The severity sweep shows that the internal disparity, and with it the detection performance, decreases smoothly as the training distribution approaches balance. The results highlight the importance of methods that provide deeper insight into the behavior of AI models.

\end{abstract}




\end{frontmatter}


\section{Introduction}
\label{sec:intro}

Artificial Intelligence (AI), particularly through the development of neural networks and machine learning algorithms, has catalyzed transformative changes across diverse domains. From optimizing logistics to predicting market trends, AI's ability to process massive volumes of data and find overlooked patterns has led to more accurate predictions, quicker responses, and informed decision-making. However, the integration of AI into critical decision-making processes also brings significant ethical and practical challenges to the forefront, especially with respect to fairness, transparency, and accountability.

The sphere of influence of AI algorithms spans a wide spectrum of domains, from the financial sector to the criminal justice system. In finance, AI-driven algorithms have revolutionized risk assessment, enabling lenders to make informed decisions about creditworthiness with unprecedented accuracy. Similarly, in law enforcement, the ability of AI to analyze patterns and anomalies has taken predictive policing and resource allocation to new levels of efficiency \citep{stone2016artificial}.

However, as AI's role in shaping decisions expands, a number of ethical complexities also unfold. A fundamental challenge lies in ensuring that the algorithms that guide decision-making are grounded in fairness, transparency, and accountability. The opacity of deep neural networks can render decision processes inscrutable, with outcomes seemingly coming from a ``black box'' \citep{szegedy2014intriguing}. This lack of transparency introduces a layer of unpredictability, potentially eroding trust and understanding, posing a challenge for practical AI applications.

Furthermore, algorithms can unintentionally perpetuate biases present in the data on which they are trained, leading to decisions that reflect and even amplify societal inequalities. Cases of algorithmic discrimination, from biased hiring practices to discriminatory criminal sentencing, have highlighted the urgent need to address these issues. The EU AI Act \footnote{\url{https://artificialintelligenceact.eu/}}, which is a landmark regulatory framework designed to ensure the safe and ethical deployment of AI technologies, recognizes the profound implications for fundamental rights and privacy.


In light of these critical issues, this work aims to explore the impact of bias on the learning process of computer vision algorithms, particularly face biometric systems, placed at the top of the list of High-Risk AI Systems of the EU AI Act, due to the personal information they process, such as identity, gender, ethnicity, and age.

The existing literature on bias analysis is mainly focused on the inputs $\mathcal{X}$ \citep{tommasi2017bias,zhang2018examining,wang2020bias} and the outputs $O$ \citep{buolamwini2018GenderShades,zisserman2018BlindEye,serna2020discrimination}.  This study goes beyond proposals that seek to model bias through performance measures \citep{2023_COMPSAC_BiasAI_N-sigma_DeAlcala,SOLANO2026113616}. The exclusive focus on performance is insufficient to identify and comprehend the origins of bias, hindering efforts to mitigate it effectively. Traditional output-based evaluations treat the model as a ``black box'', failing to capture \textit{how} and \textit{where} bias is encoded during the learning process.

To address this gap, the primary objective of this research is to propose a multi-level framework that investigates the internal mechanisms of Convolutional Neural Networks (CNNs) to uncover the origins of demographic bias. Specifically, our aim is to address the following research questions.

\begin{enumerate}
    \item How do data biases alter the distribution of the learned latent space?
    \item In what ways do neuron activation levels correlate with class under-representation?
    \item Can bias be systematically detected directly from the learned parameters of convolutional filters, completely independent of input data?
\end{enumerate}

By adopting a holistic view of the model at all possible levels --geometry of embeddings, neuron activations, and convolutional filters (or kernels) -- we aim to better understand biased learning (see Fig. \ref{fig:taxonomy}). The key contributions of this work are summarized as follows:

\begin{itemize}[noitemsep,topsep=0pt]
    \item \textbf{A comprehensive taxonomy of bias effects in the learning process:} We present a multi-level taxonomy outlining how bias propagates through the data-driven learning process of CNNs. We (i) show that the learned latent space is less effectively distributed in biased models, (ii) reveal latent correlations between demographic bias and activation levels, and (iii) demonstrate that networks encode bias-related patterns within their convolutional filter parameters.
    \item \textbf{Three novel bias detection methods:} Based on our taxonomy, we propose three specific bias detection approaches: (i) SpaceBias (Level 2), which quantitatively assesses bias based on the representation of the learned latent space; (ii) ActivationBias (Level 3), an efficient method to detect bias by examining activation levels; (iii) WeightBias (Level 4) a method to detect bias directly from the model's weights. 
    \item \textbf{Extensive experimentation in face biometrics and MNIST:} We provide reproducible experiments using the public DiveFace \citep{SensitiveNets2021} and MNIST \citep{mnist98} datasets. To validate our Level 4 methodology, we trained and evaluated more than 127,000 models with varying degrees of bias.

\end{itemize}

\subsection*{Delineation from prior conference work}
\label{sec:delineation}
\begin{figure}[t!]
    \centering
    \includegraphics[width=.95\columnwidth]{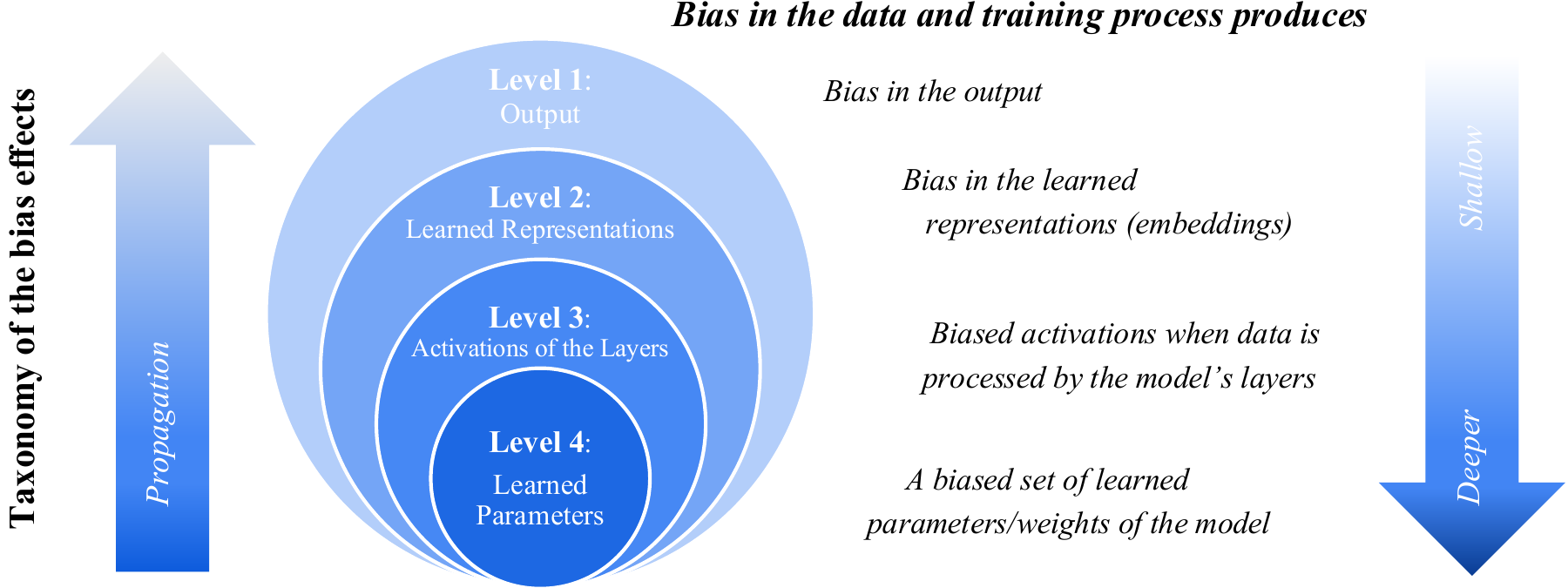}
    \caption[]{Taxonomy of the effects of bias in the learning process of neural networks.}
    \label{fig:taxonomy}
\end{figure}

Preliminary versions of the subsets of this study were presented at IAPR ICPR 2021 \citep{serna2020insidebias} (InsideBias, the Level-3 activation analysis) and at the AAAI SafeAI Workshop 2021 \citep{serna2021IFBiD} (IFBiD, the Level-4 weight-based detector). The journal version differs from those conference papers along three axes:
\begin{enumerate}[noitemsep,topsep=2pt]
\item \textbf{A new method (entirely new in this paper),} \emph{SpaceBias}, the Level-2 latent-space analysis based on neighbor-probability distributions and the Kolmogorov--Smirnov test, was not part of either conference papers.
\item \textbf{Extension of InsideBias.} \emph{ActivationBias} extends InsideBias with deeper analyses of activations and a Mann–Whitney U test for interpretable detection.
\item \textbf{A unified multi-level framework with MNIST experiments.} The most important new contribution is the integration of all three levels into a single taxonomy (Fig.~\ref{fig:taxonomy}) with consistent notation, dataset, and protocol.
\end{enumerate}

The remainder of the paper is structured as follows: Section 2 summarizes previous work. Section 3 presents our methodology. Section 4 describes the evaluation procedure in Face Biometrics and MNIST. Section 5 presents the experimental results in Face Biometrics and MNIST. Finally, Section 6 summarizes the main conclusions.

\section{Previous Work}
\label{sec:previouswork}

Bias in computer-vision models has been studied predominantly from an input--output perspective that treats the network as a black box. \citet{stock2018bias} probe a model with a hand-selected set of adversarial examples to decide whether it is biased, extending the example-based critique of \citet{kim2016interpretability}, while \citet{schaaf2021towards} compare attribution maps (Grad-CAM, Score-CAM, Integrated Gradients, LRP-$\epsilon$) as bias indicators and report that they disagree and occasionally mislead. Such analyses establish \emph{that} a model is biased but reveal little about where the bias resides. In contrast, the four levels of our taxonomy (Fig.~\ref{fig:taxonomy}) conceptualize bias as a phenomenon that progressively emerges and becomes encoded throughout the neural network hierarchy. We organize the discussion around these four levels, contrasting each with its closest prior work.

\textbf{Level~1 (output).} Most fairness research reads bias from prediction disparities and group metrics. \citet{ntoutsi2020bias} survey the problem at large; \citet{mehrabi2021survey} systematize fairness definitions and \citet{caton2024fairness} survey how they are measured, while \citet{hort2023bias} review mitigation across pre-, in-, and post-processing. In face biometrics, \citet{terhorst2021biases} document recognition biases beyond demographics, and \citet{SOLANO2026113616} propose an equity index tailored to operational face systems. Most remedies act on the data: \citet{DEANDRESTAME2025103099} balance training with synthetic faces, and \citet{perona2023benchmarking} construct a causal benchmark that manipulates input attributes. These input/output remedies say little about how bias is represented internally---a methodological gap that \citet{gluge2020not} make explicit. Level~1 is thus our established baseline, and we move inward.

\textbf{Level~2 (learned representations).} Prior work tends to \emph{shape} the embedding rather than read it: \citet{creager2019flexibly} learn fair representations by disentanglement, and \citet{wang2019balanced} show that even balanced data leave gender bias in image embeddings. Earlier attempts to read bias from representations either remove it adversarially \citep{zisserman2018BlindEye} or examine CNN features against dataset bias \citep{zhang2018examining}, typically qualitatively. SpaceBias instead makes the latent geometry quantitative and training-free, characterizing the penultimate space with neighbor-probability distributions and a Kolmogorov--Smirnov test.

\textbf{Level~3 (layer activations).} Intermediate activations have drawn comparatively little attention as a bias signal, even as evidence accumulates that networks lean on shortcuts: \citet{geirhos2020shortcut} frame shortcut learning in general terms, \citet{sagawa2020investigation} show that overparameterization amplifies spurious correlations, and \citet{neuhaus2023spurious} detect such features at ImageNet scale. ActivationBias targets this regime directly, providing a dynamic characterization of how strongly different groups activate the deep, task-specific layers and testing the gap with a Mann--Whitney $U$ statistic.

\textbf{Level~4 (learned parameters).} \citet{kirichenko2023last} show that networks internalize dataset shortcuts in their weights, so that retraining only the last layer can undo a bias the features still encode. WeightBias pushes this further: rather than retrain, it recovers the \emph{type} of bias from the convolutional filters alone, with no access to inputs or outputs.

Across the four levels, our contribution is to move from \emph{whether} biased outcomes occur to \emph{where}, \emph{when}, and \emph{how} bias forms and is stored inside the network.

\section{Methodology: Analyzing Bias at Multiple Levels}
\label{sec:methodology}
\begin{figure}[t!]
    \centering
    \includegraphics[width=.95\columnwidth]{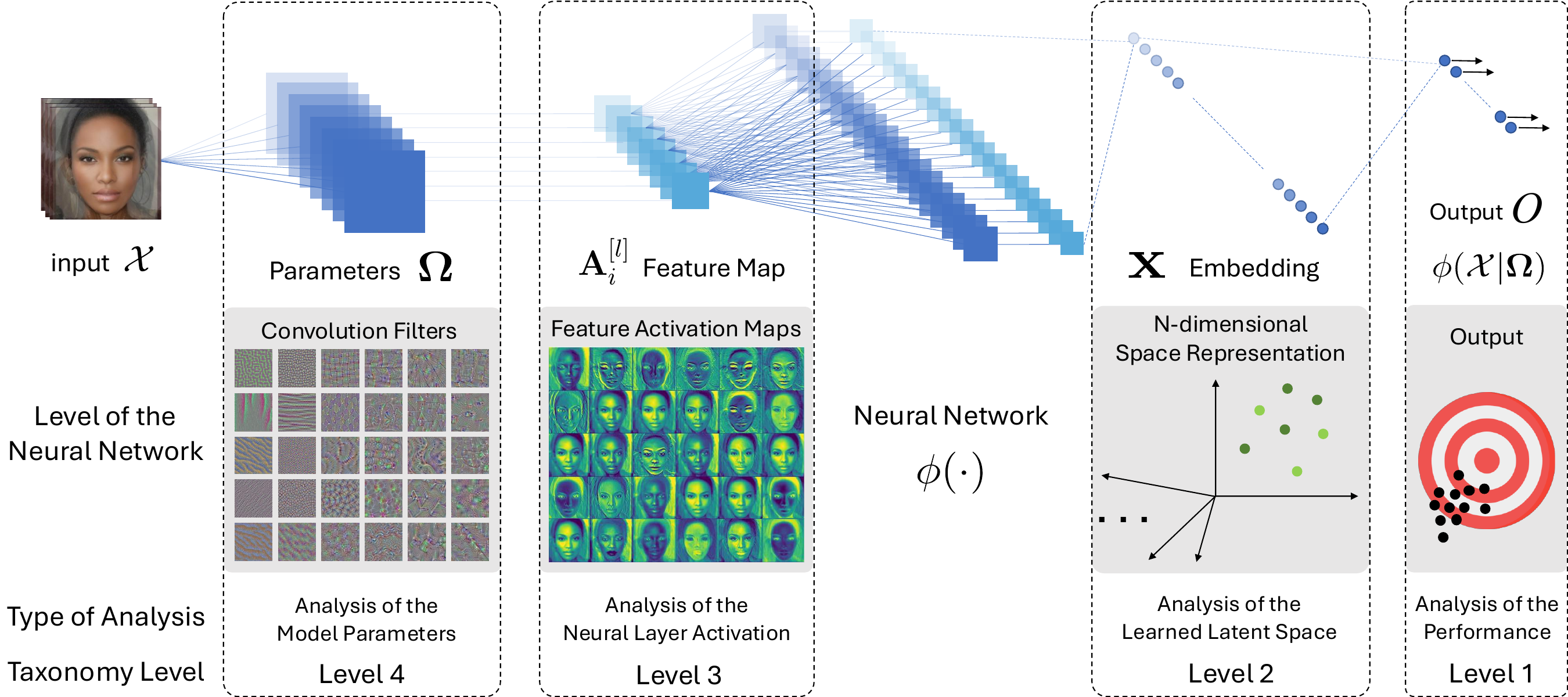}
    \caption[]{\textbf{Methodology diagram} for analyzing bias at 4 levels: 1) Model Performance; 2) Latent Learned Space; 3) Layer Activations; and 4) Model Parameters.}
    \label{fig:methodology}
\end{figure}

In this section, we outline a systematic methodology (illustrated in Figure \ref{fig:methodology}) for exploring the intricacies of neural network architectures, focusing on how biases manifest across various levels of the network, from embeddings and activations to convolutional filters.

\textbf{Bias at the input-output level} (Level 1) is the most commonly studied effect. This level captures how biases in the distribution of input data propagate through the network and influence the output of the model. We do not focus on it in this work; we go beyond the output to analyze the hidden structures and mechanisms that influence bias. 

We begin our analysis with \textbf{embeddings} (Level 2) — the rich, condensed representations learned by the network. These embeddings often encapsulate biases present in training data, providing fertile ground for detecting bias. By dissecting these embeddings, we aim to understand how biases are represented in the latent space, where data are compressed into a more abstract form. This analysis is closely related to Input-Output bias, but focuses on understanding how biases influence the internal representations of the network rather than the final output.

Moving beyond the latent space, we investigate \textbf{layer activations}\footnote{We refer to activation as the output of the activation function after each convolution.} (Level 3). This approach examines how the network responds to different inputs at each layer, allowing us to observe how these responses vary across different sets of data.

At the deepest level, we explore \textbf{bias in the weights} of the network (Level 4). Unlike the previous levels, this analysis focuses solely on the learned convolutional filters, disregarding the inputs and outputs. This level of analysis aims to demonstrate how bias is embedded even in the most fundamental parameters of the network, as they play a crucial role in shaping the network's understanding and representation of features.

\subsection{Detecting Bias in the Learned Latent Space (Level 2)}
\label{sec:level2}

The mathematical description of what a neural network does in each layer is to transform elements in the input space into a target space by means of a function. This mapping is learned during the training of the neural network, where the model adjusts the weights to optimize performance on a specific task. The space in which each layer of the network transforms is called the latent space, the latent feature space, or the embedding space. Elements in this space (\textit{embeddings}) capture the essential semantic relationships between entities, enabling efficient downstream tasks such as classification. By analyzing the relationships between embeddings, we can identify patterns that may indicate underlying biases.

A model is biased with respect to a specific variable when it behaves differently depending on the variable's value. We posit that a biased latent space is one in which a particular group of data or class is poorly distributed within it. We focus on the space of the layer prior to the classification layer. This is the representation space to which the application of a linear transformation generates the classification probabilities. This means that if this space is not well modeled, probably neither is the model's output.

Elements in this representation space are $n$-dimensional vectors, which can be interpreted geometrically as single points in $\mathbb{R}^n$. Direct visualizations in three dimensions will be a reduction in which valuable geometric information may be lost. Therefore, to study in low dimensions how points are distributed in this $n$-dimensional space, we use neighbor probabilities, which are the basis of the popular Stochastic Neighbor Embedding \cite{hinton2002SNE} and Neighborhood Components Analysis \citep{hinton2004NCA}.

The probability that point $i$ chooses $j$ as its neighbor is:
\begin{equation}
    \label{eqn:pij}
    p_{ij}= \frac{\text{exp}(-||\textbf{x}_i-\textbf{x}_j||^2)}{\sum_{k\neq i} \text{exp}(-||\textbf{x}_i-\textbf{x}_k||^2)}
\end{equation}

In this equation, \(\textbf{x}_i\) and \(\textbf{x}_j\) represent the points \(i\) and \(j\) as vectors. The numerator represents the affinity between the points \(i\) and \(j\), while the denominator is a normalization term that ensures the probabilities sum to one over all possible neighbors \(k \neq i\).

Under this stochastic selection rule, we can compute the probability \(p_i\) that point \(i\) will be correctly classified within a set of \(N\) neighbors.  Let \(C_i = \{j \in \mathcal{N} \mid c_i = c_j\}\) denote the set of points in the neighborhood \(\mathcal{N}\)  that share the same class as $i$. Then:

\begin{equation}
    \label{eqn:pi}
    p_i= \sum_{j \in C_i} p_{ij}
\end{equation}

\begin{figure}
\includegraphics[width=\textwidth]{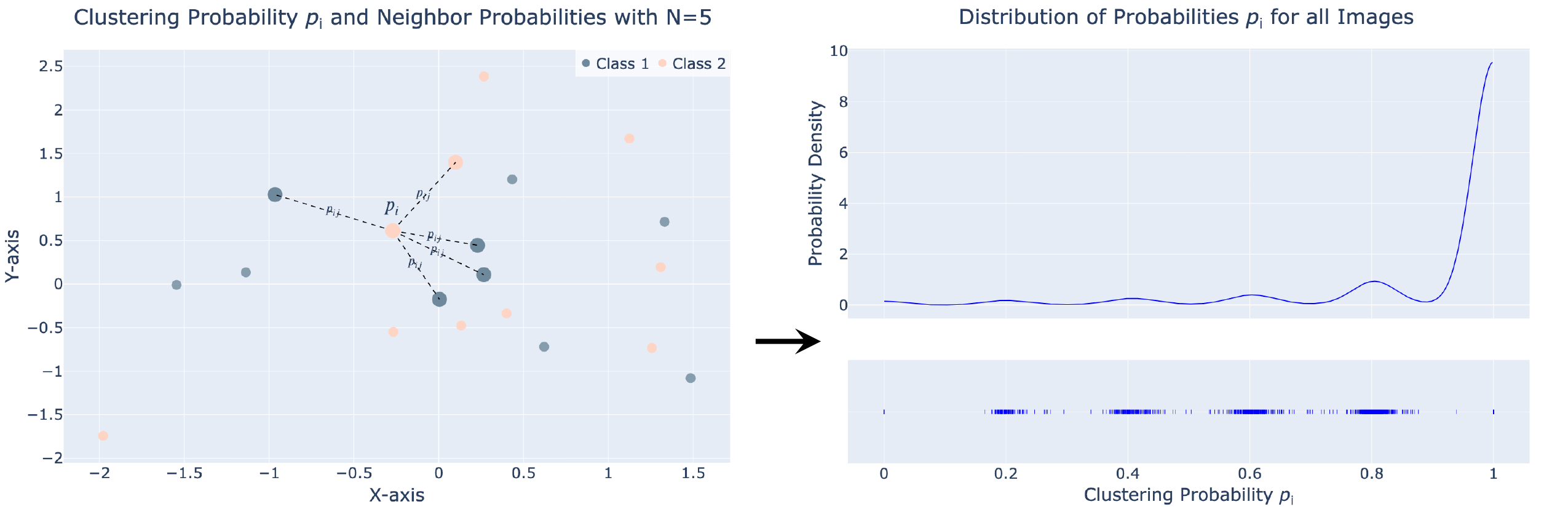}
\caption[]{\textbf{Measuring the Learned Latent Space.} 1$^{st}$: \textbf{Compute Clustering Probability}. For each sample, calculate the clustering probability $p_i$ considering $N$ neighbors. The probability $p_i$ of a sample represents the likelihood that neighboring samples in the latent space belong to the same class (e.g. female or male) as the sample under consideration. 2$^{nd}$: \textbf{Analyze Probability Distribution}. 
Determine the distribution of clustering probabilities $p_i$ for all samples to assess the overall classification consistency within the latent space. The right figure shows the distribution of probabilities with $N$=5. }
\label{fig:latent_distribution}
\end{figure}

\noindent Here, \(p_i\) is the sum of the neighbor probabilities \(p_{ij}\) for all neighbors \(j\) that belong to the same class as \(i\). This sum represents the probability that point \(i\) will be correctly classified based on its proximity to other points of the same class within the neighborhood. By measuring the probability of all points belonging to their respective classes, we obtain a probability distribution. This probability distribution provides an approximation of how the $n$-dimensional space is structured over a distance of \textit{N} neighbors. Figure \ref{fig:latent_distribution} illustrates this distribution of probabilities $p_i$.

The fact that the probability distributions of different groups are diverse indicates that the layout of the latent space for these groups is dissimilar. The sense in which these spaces are dissimilar is that some groups have the samples of both classes more intermingled than the samples of other groups. The direct consequence is that it is more difficult for a classifier (final dense layer) to separate between the two classes. Since the final classification layer operates on these embeddings, separability there bounds what that head can achieve. The real head is a linear softmax, so $p_i$ measures $NN$-separability, which is tight for our method and strongly indicative (not identical) for the linear classifier. 

To quantitatively compare these probability distributions, we use the the two-sample Kolmogorov--Smirnov ($KS$) test. The $KS$ statistic, is defined as the maximum vertical distance between the empirical cumulative distribution functions (eCDFs) of two samples, denoted by $F_{1,n_1}(z)$ and $F_{2,n_2}(z)$. The $KS$ test is a statistical method employed to assess the similarity of the cumulative distributions of two sets.  It is a non-parametric test, which means that it does not make any assumptions about the underlying distribution of the data. It tests the hypotheses $H_0 : F_1 = F_2$ versus $H_1 : F_1 \neq F_2$, where $F_n$ denotes the empirical cumulative distribution function (eCDF) for $n$ independent and identically distributed (i.i.d.) observations.  Essentially, it determines whether the differences between the two probability distributions are significant or not.

The $KS$ statistic is given by:

\begin{equation}
    \label{eqn:Dn}
    D_n= \sup_{z} \big |F_{1,n_1}(z) - F_{2,n_2}(z)\big |
\end{equation}

The significance level $\alpha$ represents the probability of making a Type I error, which occurs when we incorrectly reject a true null hypothesis. In other words, it quantifies the risk of concluding that there is a significant difference between two samples when, in reality, there is none. A common choice for the significance level is $\alpha = 0.05$, corresponding to a 95\% confidence level. This means that if we were to perform the same experiment many times, approximately $95$\% of the time, we would correctly fail to reject the null hypothesis when it is true. In our experiments, we use the slightly more conservative $\alpha = 0.01$ ($99$\% confidence) because the KS test is high-powered at our sample sizes ($n_1, n_2 = 6{,}000$ per group).

For large samples, the null hypothesis ($H_0$) is rejected with confidence level $\alpha$ if:

\begin{equation}
    \label{eqn:null}
    D_n > \sqrt{-\text{ln}(\frac{\alpha}{2}) \cdot\frac{n_1+n_2}{2n_1n_2}}
\end{equation}

\paragraph{Why the $KS$ test}
The clustering probabilities generated in the latent space are highly non-normal, heavily skewed, and multimodal. Therefore, parametric tests that assume normality (such as t-tests or ANOVA) are mathematically invalid for this context. We selected the two-sample $KS$ test because it is a non-parametric test that makes no assumptions about the underlying data distribution beyond continuity, which is appropriate for probabilities $p_i \in [0,1]$. Furthermore, because bias in the latent space alters the concentration of probability mass rather than simply shifting a unified mean, the $KS$ test's sensitivity to both the location and the shape of the eCDF makes it the ideal metric to quantify these specific structural discrepancies. Alternatives such as the Anderson–Darling test were not chosen because they overweight tail differences, which are less relevant when $p_i$ lies in a bounded interval. The Wasserstein-1 distance was also considered, but the $KS$ test provides a closed-form significance threshold that is more interpretable in a multi-comparison setting.

\paragraph{Definition 1 (Latent-space bias)}
Let $\phi:\mathcal{X} \to \mathbb{R}^d$ be a neural network mapping inputs to the penultimate latent space, and let $\mathcal{V}$ be a variable (e.g. categorical attribute). For each input $\textbf{x}_i \in \mathcal{X}$ with neighborhood size $N$, let $p_i(\textbf{x}_i) \in [0,1]$ denote the clustering probability defined in Eq.~\ref{eqn:pi}. Let $F_v$ denote the empirical cumulative distribution function (eCDF) of $\{p_i(\textbf{x}) : \textbf{x} \in \mathcal{V} = v\}$. We say that $\phi$ is \emph{biased in the latent space} with respect to $\mathcal{V}$ at confidence level $\alpha$ if there exists $v_1, v_2$ such that the Kolmogorov--Smirnov test rejects $H_0: F_{v_1} = F_{v_2}$ at level $\alpha$ (Eq.~\ref{eqn:null}). The magnitude of the latent-space bias indicator is:
\begin{equation}
\label{eqn:LBI}
\Gamma_v(\phi; v_1, v_2) \;=\; D_n\bigl(F_{v_1},\, F_{v_2}\bigr) \;\in\; [0, 1].
\end{equation}
Larger $\Gamma_v$ indicates a more strongly biased latent geometry; $\Gamma_v = 0$ corresponds to identically distributed clustering probabilities, i.e.\ no detectable latent-space bias under a sample of $n_1 + n_2$ observations.

\subsection{Detecting Bias in the Activations of the Layers (Level 3)}

Convolutional Neural Networks are composed of a large number of stacked filters. These filters are trained to extract the richest information for a predefined task. As the input, for example, an image, transverses the network, the filters are activated. Without loss of generality, we present a method to establish a relationship between activation and bias for Convolutional Neural Networks (CNNs). Similar ideas can be extended to other neural learning architectures. 

In a convolutional layer of a CNN, the feature maps of the previous layer are convolved with the filters (also known as kernels) and put through the activation function to form the output feature map. The output $\textbf{A}^{[l]}$ of layer $l$ consists of $m^{[l]}$ feature maps of size $n_1^{[l]} \times n_2^{[l]}$, where $m^{[l]}$ is the number of filters in layer $l$. The $i^{\textnormal{th}}$ feature map of layer $l$, denoted as $\textbf{A}_i^{[l]}$ is computed as: 

\begin{equation}
\label{eqn:feature_map}
    \textbf{A}_i^{[l]} = g^{[l]}\left(\sum_{j=1}^{m^{[l-1]}} \textbf{f}_{ij}^{[l]} * \textbf{A}_j^{[l-1]} + \textbf{b}_i^{[l]}\right)
\end{equation}

\noindent where $g^{[l]}$ denotes the activation function of layer $l$, $*$ is the convolutional operator, $\textbf{b}^{[l]}_i$ is a bias vector in layer $l$ for the $i^{\textnormal{th}}$ feature map and $\textbf{f}_{ij}^{[l]}$ is the filter connecting the $j^{\textnormal{th}}$ feature map in layer $(l-1)$ with the $i^{\textnormal{th}}$ feature map in layer $l$. The average activation of the  $i^{\textnormal{th}}$ feature map at layer $l$ is calculated as:

\begin{equation}
\label{eqn:average}
    \overline{A_i^{[l]}} = \frac{1}{n_1^{[l]} \cdot n_2^{[l]}} \sum_{h=1}^{n_1^{[l]}} \sum_{w=1}^{n_2^{[l]}} A_{i}^{[l]} (h,w)
\end{equation}

\noindent where $(h,w)$ are the spatial coordinates of the output $A_{i}^{[l]}$. The activation, $\hat{\lambda}^{[l]}$, is calculated as the maximum of $\overline{A_i^{[l]}}$ for all feature maps in the layer $l$: 

\begin{equation}
\label{eqn:activation}
    \lambda^{[l]} = \max_{i}\left(\overline{A_i^{[l]}}\right)
\end{equation}

We have evaluated both the average and the maximum, but the maximum resulted in a better estimator. Our intuition is that the maximum works best because it is exactly the mechanism used in the MaxPooling layers of the network.

We then compare per-group distributions of the activation $\lambda^{[l]}$ (Eq.~\ref{eqn:activation}) for every possible set of samples $\mathcal{D}_{v_k}$ of a variable's values $v_k$. For each pair of groups $v_1, v_2$ we apply the two-sample Mann--Whitney $U$ test and report the \emph{probability of superiority}:
\begin{equation}
    \label{eqn:theta}
    \hat\theta_{v_1 v_2} \;=\; \frac{U_{v_1 v_2}}{n_{v_1} n_{v_2}}
    \;=\; \mathbb{P} \bigl(\lambda^{[l]}(\mathcal{D}_{v_1}) > \lambda^{[l]}(\mathcal{D}_{v_2})\bigr),
    \qquad \mathcal{D}_{v_1} \sim (\mathcal{V}{=}v_1),\; \mathcal{D}_{v_2} \sim (\mathcal{V}{=}v_2),
\end{equation}
which is the probability that a randomly drawn sample from group $v_1$ activates more strongly than one from group $v_2$. A value of $\tfrac12$ indicates stochastically equal activations; deviations towards $0$ or $1$ indicate that one group consistently activates less or more. 

\paragraph{Definition 2 (Activation bias)}
Let $\phi$ be a convolutional neural network with $L$ convolutional layers, and let $\mathcal{V}$ be a variable (e.g., categorical attribute). For a pair of values $v_1, v_2$, let $\hat\theta_{v_1 v_2}$ be the probability of superiority, estimated by the normalized Mann--Whitney $U$ statistic (Eq.~\ref{eqn:theta}). We say that $\phi$ is \emph{biased in the activations} at layer $l$ with respect to $\mathcal{V}$ if there exist $v_1, v_2$ such that the Mann--Whitney test rejects $H_0: \theta_{v_1 v_2} = \tfrac12$. The magnitude of the \textit{Activation Bias Indicator} is
\begin{equation}
\label{eqn:ABI}
\Lambda^{[l]}_v \;=\; \bigl|1- 2\,\hat\theta_{v_1 v_2}\bigr| \;\in\; [0, 1],
\end{equation}
which is the absolute rank-biserial correlation between the two groups' activations. Larger $\Lambda^{[l]}_v$ indicates more strongly separated activation distributions; $\Lambda^{[l]}_v = 0$ corresponds to stochastically equal activations ($\hat\theta_{v_1 v_2} = \tfrac12$), i.e., there is no detectable activation bias at layer $l$, while the sign of $(\hat\theta_{v_1 v_2} - \tfrac12)$ identifies the favored group.

\subsection{Detecting Bias in Parameters of the Convolution Filters (Level 4)}

We analyze the parameters $\mathbf{\Omega}$ of Convolutional Neural Networks $\phi(\cdot)$. The aim is to find patterns in the convolutional filters of the model associated with biased outcomes. The purpose of convolution is to detect the filter pattern in the image, producing a high signal when present and a low signal when not present. Following that reasoning, a high activation signal indicates that the image contains the pattern (that is why stronger activations are usually related to the detection of highly discriminative features \cite{zeiler2014visualizing}). Therefore, when a model has not successfully learned the patterns associated with certain classes, these patterns are not detected (activated) by the network, which leads to poor performance on those classes. This outlines the connection between the failure to activate certain patterns and poor performance (bias).

The training process of Neural Networks is usually not deterministic and the resulting parameters $\mathbf{\Omega}$ depend on several elements: training data, learning architecture (e.g., number of layers, number of neurons per layer, etc.), training hyper-parameters (e.g., loss function, number of epochs, batch size, learning rate, etc.), initialization parameters, and optimization algorithm.

Due to the non-deterministic nature of the network training process $\phi(\cdot |\mathbf{\Omega})$, the same training subset is likely to give rise to different $\mathbf{\Omega}$. The reason for this is that since the solution space is very large, the solution (which is iteratively approximated) typically arrives at a local minimum that depends on the initialization, the particular training configuration and the order of the data \cite{lecun2015deep}. In CNNs, this translates into the fact that no two networks are identical in terms of the learned convolutional filters. The challenge, therefore, is to be able to detect patterns in the filters in different positions and configurations. The problem is analogous to detecting patterns in images, where one can be in different parts of an image. 

Our proposed approach involves allowing another neural network to uncover bias patterns. Supervised neural network systems are particularly useful for this endeavor as their objective is to identify correlations between input patterns and output. We designed the biased pattern detector as a Neural Network $\psi(\cdot)$ represented by its parameters $\mathbf{\Theta}$, concisely described as $\psi(\cdot | \mathbf{\Theta})$. In this approach, we train the pattern detector using a dataset of biased and unbiased models $\phi(\cdot)$.  These models $\phi(\cdot |\mathbf{\Omega})$ are trained on data related to a specific task $T$, such as image classification or face recognition, learning the parameters $\mathbf{\Omega}$ that optimize their performance on this task. They are trained with subsets $\mathcal{S}_1, ..., \mathcal{S}_n \subset \mathcal{D}$  from the entire database $\mathcal{D}$. Each subset $\mathcal{S}_i$ is intentionally biased in some way. For example, a subset might over-represent or under-represent certain classes or demographic groups. This means that each model $\phi(\cdot)$ is trained using a different subset $\mathcal{S}_i$ of the entire dataset $\mathcal{D}$, which introduces specific biases embedded in its learned parameters $\mathbf{\Omega}$. 

Training $\psi(\cdot)$ on the parameters $\mathbf{\Omega}$, it learns to identify if certain biases are present in the parameters themselves. This setup allows the detector to identify biased patterns by analyzing the learned parameters of task-specific models, rather than directly analyzing the task outcomes. 

Designing the detector $\psi(\cdot)$ involves exploring various neural network architectures. The choice of architecture requires careful consideration of how well it can process the input parameters $\mathbf{\Omega}$ from the models $\phi(\cdot |\mathbf{\Omega})$.  The effectiveness of the detector  depends on how well it is tailored to handle these input parameters, and its design must be informed by the characteristics of the parameters $\mathbf{\Omega}$. (See \ref{appx:detector} for more details.)

\section{Experimental Set-up: Two Applications}
\label{sec:setup}

We instantiate the methodology of \S\ref{sec:methodology} in two computer vision applications, chosen so that they vary the task, the data domain, and the bias mechanism while keeping the detection methodology constant:
\begin{itemize}[noitemsep,topsep=2pt]
\item \emph{Application 1 (\S\ref{subsec:app_face}): Face biometrics.} Gender classification of face images, with demographic under-representation as the bias mechanism. The bias variable is ethnicity (Asian, African/Indian, Caucasian).
\item \emph{Application 2 (\S\ref{subsec:app_mnist}): Colored MNIST.} Digit classification of synthetic colored-MNIST images, with a skewed color distribution as the bias mechanism. The bias variable is the color of the image (red, green, or blue).
\end{itemize}

The two applications play complementary roles. The ResNet face models are our analysis instrument: a small, controllable set, deep enough to make the effect interpretable, on which we study \emph{how} bias reshapes the latent space (\textit{Level~2}) and the activations (\textit{Level~3}). \emph{Detection} is then validated at scale on two independent model populations ---the tiny face models and the colored-MNIST models--- each containing thousands of biased and unbiased networks with known bias labels. Running all three detectors on the MNIST model population lets us measure detection accuracy and, because the two domains differ in task, data, and bias mechanism, assess how much of the methodology is general rather than face-specific. Colored MNIST plays one further role that faces cannot afford: because it is inexpensive to train, we use it to sweep the \emph{severity} of the bias and measure how detection degrades as the training distribution moves from heavily skewed toward balanced.

The two roles also reflect how the detectors work. The \textit{Level~2--3} detectors are training-free statistical tests applied to a single model, whereas the \textit{Level~4} detector is a learned model trained on a population. Because per-model tests are run individually and this is computationally costly, \textit{Level~2--3} detection experiments use a fixed subsample of $100$ models per configuration in each domain, which is sufficient for a stable accuracy estimate; the \textit{Level~4} detector is trained on the full populations.

\subsection{Application 1: Face Biometrics}
\label{subsec:app_face}

We train neural networks for gender classification from face images. Since establishing ground-truth genetic sex is not possible, we use gender as a proxy for sex, as a simplified instance of a real-world problem.

\subsubsection{Database}
\label{subsec:database}

We use the DiveFace dataset \citep{SensitiveNets2021}, a collection of facial images categorized into six classes by gender and ethnicity. Ethnicity attributes include not only skin color, but also more complex anthropomorphic facial features. The dataset contains 72,000 images: three images for each of 24,000 identities, balanced at 12,000 per gender and ethnicity. Users are grouped by gender (male or female) and by three categories related to ethnic physical characteristics: \textit{Asian}, \textit{African / Indian}, and \textit{Caucasian}.

In all protocols below, models are trained \emph{from scratch} with Glorot-uniform initialization \citep{glorot2010init}; no ImageNet \citep{deng2009imagenet} or VGGFace \citep{cao2018vgg2} pretraining is used, so that observed biases can be unambiguously attributed to the DiveFace training distribution. Training and test images are equally divided between males and females.

\subsubsection{ResNet Face Models: Analysis of the Latent Space and Activations (Levels 2--3)}
\label{subsec:ResNet}

For the analysis of the latent space (\textit{Level~2}) and the activations (\textit{Level~3}), we use a compact ResNet architecture \citep{he2016resnet}, based on residual connections -- a representative baseline for facial attribute detection \citep{ranjan2018faces}. It comprises over 370K parameters in 22 convolutional layers; the layer-by-layer specifications are given in \ref{appx:architecture}.

We train five models that differ only in the demographic distribution of their training data, yielding five ResNet face models in total. Table~\ref{tab:compact_models} reports the per-group gender-classification accuracy along with the training protocol. The database was split 75\% / 25\% into training and test sets; the 18,000-image test set is distributed equally across the three ethnic groups and none of the test users appear in the training set. Both \textit{biased} and \textit{unbiased} models are trained with the same number of samples (18,000); only the \textit{Unbiased~(+)} model is trained with three times more data (54,000). Each biased model over-represents one group (90\% of the images), producing the skewed per-group performance visible in the table.

\begin{table*}[t]
    \centering
    \caption{Joint overview of the dataset training configurations and model performance of ResNet models. Number of images per demographic group (Asian, African/Indian, Caucasian) utilized across the biased and unbiased training protocols. Gender classification accuracy (\%) across the demographic groups for the ResNet architecture. The dataset split was 75\% training and 25\% test.}
    \label{tab:compact_models}
    \small
    \setlength{\tabcolsep}{4pt}
    \resizebox{0.85\textwidth}{!}{
        \begin{tabular}{l cccc ccc}
            \toprule
            & \multicolumn{3}{c}{\textbf{Training Samples}}& & \multicolumn{3}{c}{\textbf{ResNet Accuracy (\%)}} \\
            \cmidrule(r){2-4} \cmidrule(lr){6-8}
            \textbf{Model}& \textbf{Asian} & \textbf{Afr./Indian} & \textbf{Cauc.} & & \textbf{Asian} & \textbf{Afr./Indian} & \textbf{Cauc.} \\
            \midrule
            \textit{Asian Biased} & 16,200& 900   & 900   & & \textbf{96.84} & 94.14 & 94.45 \\
            \textit{Afr./Indian Biased} & 900   & 16,200& 900   & & 93.29 & \textbf{96.86} & 95.40 \\
            \textit{Cauc. Biased} & 900   & 900   & 16,200& & 94.80 & 95.21 & \textbf{97.01} \\
            \textit{Unbiased}   & 6,000& 6,000& 6,000& & 95.50 & 95.35 & 96.11 \\
            \midrule
            \textit{Unbiased (+)}  & 18,000& 18,000& 18,000& & \textbf{97.47}    & \textbf{97.44}    & \textbf{98.17}    \\
            \bottomrule
        \end{tabular}
    }
\end{table*}

\subsubsection{Tiny Face Models: Detection in the Parameters (Level 4)}
\label{subsec:tiny}

Validating the learned weight detector of \textit{Level~4} requires a large population of task models with known bias labels. For this we use tiny face models (a standard CNN, $\approx$100K parameters; \ref{appx:architecture}), small enough to train tens of thousands of models. Using the same task and bias mechanism as the ResNet models, we train 36,000 biased models under three protocols, each over-representing one ethnic group (Table~\ref{tab:GenderWdb}). To keep detector training and evaluation independent, the dataset is split into two equal, non-overlapping halves: the bulk of the models are trained on Partition~1 and the remainder on Partition~2.

Note the gap between model sizes: the ResNet models are nearly perfect (94--98\%), whereas the tiny models range between roughly 80\% and 90\% depending on the group. This is because the architecture of the tiny models lacks the representational capacity to capture the subtle facial features needed for accurate gender classification.

\begin{table*}[t]
    \centering
    \caption{Number of models trained per partition, training data distribution for each protocol, and per-group performance (\%) for the standard tiny models on DiveFace. The dataset is split into two equal, non-overlapping halves (Partition~1 and Partition~2).}
    \label{tab:GenderWdb}
    \resizebox{1\textwidth}{!}{
        \begin{tabular}{@{}l cc ccc ccc@{}}
            \toprule
            & \multicolumn{2}{c}{\textbf{Models Trained}} & \multicolumn{3}{c}{\textbf{Training Samples}} & \multicolumn{3}{c}{\textbf{Performance (\%)}} \\
            \cmidrule(r){2-3} \cmidrule(lr){4-6} \cmidrule(l){7-9}
            \textbf{Protocol} & \textbf{Part. 1} & \textbf{Part. 2} & \textbf{Asian} & \textbf{Afr./Ind.} & \textbf{Cauc.} & \textbf{Asian} & \textbf{Afr./Ind.} & \textbf{Cauc.} \\
            \midrule
            \textit{Asian Biased}       & 10,000 & 2,000 & 12,000 & 1,000  & 1,000  & \textbf{89.5} & 81.5 & 82.9 \\
            \textit{Afr./Ind. Biased}   & 10,000 & 2,000 & 1,000  & 12,000 & 1,000  & 82.0 & \textbf{89.4} & 83.0 \\
            \textit{Cauc. Biased}       & 10,000 & 2,000 & 1,000  & 1,000  & 12,000 & 80.0 & 83.2 & \textbf{89.2} \\
            \midrule
            \textbf{Total}              & \textbf{30,000}& \textbf{6,000}& \multicolumn{6}{c}{} \\
            \bottomrule
        \end{tabular}
    }
\end{table*}

\subsection{Application 2: Colored MNIST}
\label{subsec:app_mnist}

The second application is digit classification on a controlled colored-MNIST benchmark whose bias mechanism is a skewed color distribution. Holding the methodology of \S\ref{sec:methodology} constant while changing the task (digit vs.\ gender) and the data domain (synthetic 28$\times$28 RGB vs.\ natural face crops) tests whether the multi-level framework captures something general about how CNNs encode bias. With thousands of biased and unbiased models under a single architecture, this population is our second independent detection testbed: agreement between it and the tiny face models, across domains that differ in task, data, and bias mechanism, is what lets us argue the methodology is not face-specific.

\begin{table}[t!]
    \centering
    \caption{Number of models trained per partition, training data distribution for each protocol, and performance (\%) on each color group. The MNIST training set was split into two equal, non-overlapping halves (50\% for Partition~1 and 50\% for Partition~2).}
    \label{tab:DigitWdb}
    \resizebox{1\textwidth}{!}{
        \begin{tabular}{@{}l l cc ccc ccc@{}}
            \toprule
            & & \multicolumn{2}{c}{\textbf{Models Trained}} & \multicolumn{3}{c}{\textbf{Training Samples}} & \multicolumn{3}{c}{\textbf{Performance (\%)}} \\
            \cmidrule(r){3-4} \cmidrule(lr){5-7} \cmidrule(l){8-10}
             & \textbf{Severity} & \textbf{Part. 1} & \textbf{Part. 2} & \textbf{Red}& \textbf{Green}& \textbf{Blue}& \textbf{Red}& \textbf{Green}& \textbf{Blue}\\
            \midrule
            \textbf{Red Bias} & \textbf{I} & 5,000 & 2,000 & \textbf{10,000} & 50  & 50  & \textbf{0.96} & .88 & .87 \\
             & \textbf{II} & " & " & \textbf{9,090} & 505  & 505  & \textbf{95.9} &  94.3 &  94.3\\
             & \textbf{III} & " & " & \textbf{8,080} & 1,010  & 1,010  & \textbf{95.7} &  94.8 &  94.8\\
             & \textbf{IV} & " & " & \textbf{6,060} & 2,020  & 2,020  & \textbf{95.5} &  95.0 &  95.1\\
            \textbf{Green Bias} & \textbf{I} & 5,000 & 2,000 & 50  & \textbf{10,000} & 50  & 88.4& \textbf{96.4}& 87.9\\
             & \textbf{II} & " & " & 505 & \textbf{9,090}  & 505  & 94.4 &  \textbf{96.1} &  94.4\\
             & \textbf{III} & " & " & 1,010 & \textbf{8,080}  & 1,010  & 95.0 &  \textbf{96.0} &  95.1\\
             & \textbf{IV} & " & " &  2,020 & \textbf{6,060}  & 2,020  & 95.1&  \textbf{95.5} &  95.1\\
            \textbf{Blue Bias} & \textbf{I} & 5,000 & 2,000 & 50  & 50  & \textbf{10,000} & 88.3& 88.2 & \textbf{96.4}\\
             & \textbf{II} & " & " & 505 & 505  & \textbf{9,090}  & 94.8 &  94.7 &  \textbf{96.2}\\
             & \textbf{III} & " & " & 1,010 & 1,010  & \textbf{8,080}  & 95.1 &  95.0 &  \textbf{95.9}\\
             & \textbf{IV} & " & " &  2,020 & 2,020  & \textbf{6,060}  & 95.3 &  95.1 &  \textbf{95.5}\\
            \midrule
            \textbf{Unbiased} & &5,000 & 2,000 & 3,367& 3,367& 3,367& 95.3 & 95.3 & 95.2\\
            \midrule
            \textbf{Total} & & \textbf{65,000} & \textbf{26,000} & & & &  & & \\
            \bottomrule
        \end{tabular}
    }
\end{table}

\subsubsection{Dataset construction}
\label{subsec:mnist_dataset}
We used the canonical MNIST partitions (60,000 training images and 10,000 test images) and split the training set into two equal halves of 30,000. Every image is converted to RGB and assigned one of three colors -- \emph{red}, \emph{green}, or \emph{blue} -- by coloring the \emph{background} while leaving the digit strokes white; all partitions (training partition~1, training partition~2, and the test partition) are balanced, with each color equally represented within every digit class. Bias is introduced only through the training color distribution.

\subsubsection{Architecture and training protocol}
\label{subsec:mnist_arch_protocol}

Unlike the face application, where we deliberately varied the size and type of architecture across methodologies, here we standardize on a single architecture for all three levels: since the task is simple, a small network reaches good performance and can be trained thousands of times cheaply. The colored-MNIST models use a small CNN with three convolutional layers and approximately 30$K$ parameters (\ref{appx:mnist_arch}). As in the face application, models are trained from scratch with Glorot-uniform initialization, and the dataset is split into two non-overlapping halves to keep detector training and testing independent. 

We train models under three biased families---majority \emph{red}, \emph{green}, or \emph{blue}---each at four levels of severity, from the most skewed (\textbf{I}: 99\%/0.5\%/0.5\%) to the least (\textbf{IV}: 60\%/20\%/20\%), together with an unbiased family (33\%/33\%/33\%). Table~\ref{tab:DigitWdb} lists the twelve biased configurations and the unbiased one. We train $7,000$ models per configuration, split across the two training partitions (5,000 and 2,000); the \textit{Level~2--3} detection experiments draw a $100$-models-per-configuration subsample, while the \textit{Level~4} detector is trained on the full population. The four severity levels let us report detection not as a single number but as a function of how biased the data are; for a head-to-head comparison with faces, we reference the configuration closest to the tiny-face protocol (\textbf{II}, 90\%/5\%/5\%).

\section{Findings and Interpretation}
\label{sec:results}

We present results level by level. At each level, we first \emph{analyze} how bias manifests, using ResNet face models, and then evaluate \emph{detection} at scale. At-scale statistical detectors of \textit{Levels~2--3} are validated on the colored-MNIST population, where the controlled severity sweep additionally measures how detection degrades as the training distribution moves from heavily skewed (\textbf{I}) toward balanced (\textbf{IV}). Every level pairs a disparity measure (for understanding) with a decision rule (for detection): a Kolmogorov--Smirnov (KS) test at \textit{Level~2}, the rank-biserial correlation of a Mann--Whitney $U$ test at \textit{Level~3}, and a learned classifier at \textit{Level~4}.

\subsection{Level 2: Latent Space (SpaceBias)}
\label{sec:results_l2}

The space generated by a model is high dimensional, which is why using the clustering algorithm from \cite{hinton2002SNE} can help us understand how the space is distributed. Visualization algorithms like t-SNE or PCA that project the space into three dimensions do not allow us to capture the complex relationships adequately. While with t-SNE or PCA it is observable that images from the same group are clustered together, our aim is to investigate whether the space is poorly distributed for underrepresented groups. We use nearest-neighbor probabilities to estimate the structure of the latent space, that is, the foundation algorithm of t-SNE. To quantify the difference in this structure, we use the Kolmogorov-Smirnov (KS) statistical test between the probability distributions.

\subsubsection{Analysis (ResNet face models)}
\label{subsec:res_l2_analysis}

\begin{figure}
    \includegraphics[width=\textwidth]{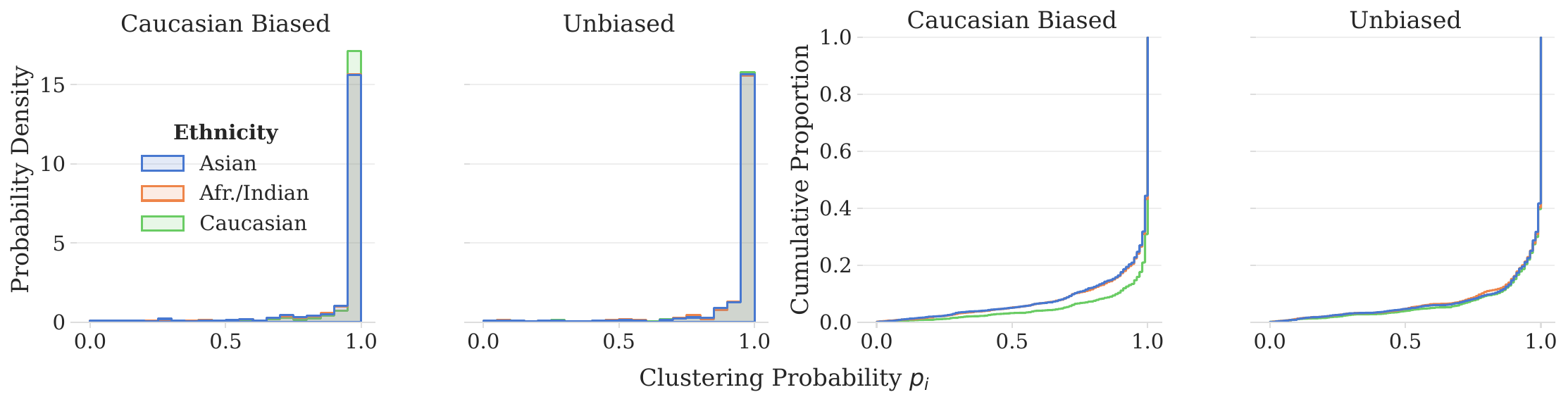}
    \caption[]{\textbf{Distribution of clustering probabilities $p_i$ for Caucasian biased and unbiased models.} The probability is computed using Eqn. \ref{eqn:pi} with neighbors N = 100. Caucasian images have higher probabilities than Asian or African/Indian images (higher spike around 1).}
    \label{fig:kde_p_face}
\end{figure}


We first inspect the distribution of clustering probabilities $p_i$ in the embedding space immediately before the classification layer. Figure~\ref{fig:kde_p_face} shows these distributions, and their empirical cumulative distribution functions (eCDF), for Caucasian-biased and unbiased ResNet models at $N=100$. In the unbiased model, the three demographic groups are almost indistinguishable, whereas in the Caucasian-biased model, the favored group (Caucasian) is clearly displaced from the other two: its $p_i$ values are systematically higher, so its eCDF separates from those of the Asian and African/Indian groups, which remain superimposed. A more concentrated distribution of $p_i$ near one indicates that an image's neighbors in the latent space more reliably share its class; the favored group enjoys this better-structured neighborhood, while the underrepresented groups do less.

\begin{figure}
\includegraphics[width=\textwidth]{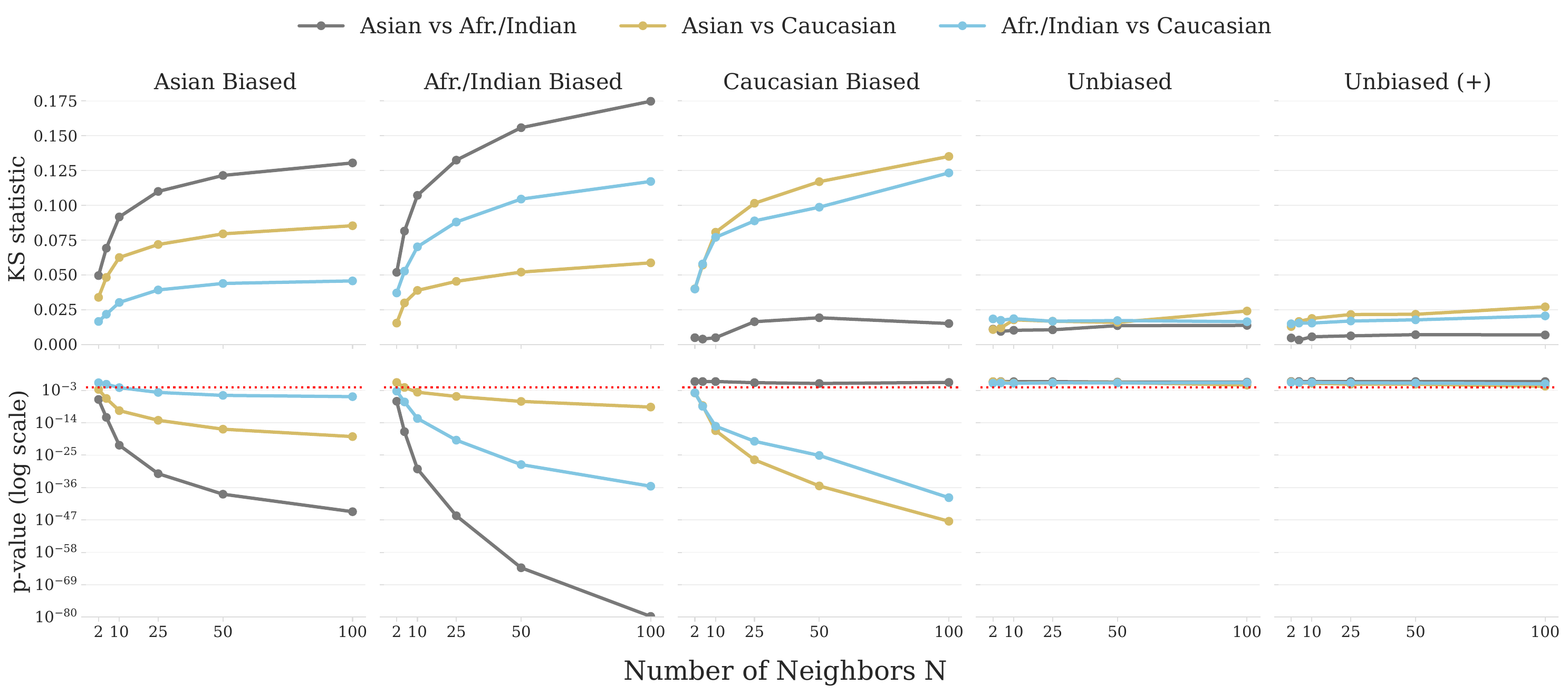}
\caption[]{\textbf{Kolmogorov–Smirnov ($KS$) test statistics and $p$-values evaluated across ResNet models.} Each column represents a specific model variant, plotting the pairwise statistics and $p$-values of probability distributions between three demographic groups against the number of neighbors $N$. The red dotted line marks the rejection threshold for the null hypothesis ($H_0$) of identical distributions with a 99\% confidence level ($\alpha =0.01$); above this line, the null hypothesis cannot be rejected. The $p$-value is in logarithmic scale for better visualization.}
\label{fig:ks_pval_N_face}
\end{figure}

To quantify these differences as a function of the neighborhood size, Figure~\ref{fig:ks_pval_N_face} reports, for each ResNet model, both the $KS$ statistic and its $p$-value for the three pairwise group comparisons as $N$ grows from $2$ to $100$. Two complementary readings emerge. The KS statistic behaves as an effect size: in each biased model the comparisons that involve the favored group grow with $N$ to $0.08$--$0.18$, while the comparison between the two non-favored groups stays small. This relationship is not symmetrical, since the inherent anthropometric differences vary depending on the specific demographic groups being compared. The Caucasian-biased model is the clearest case: the Asian-vs-African/Indian comparison is essentially null ($D_n\!\approx\!0$), confirming that the two underrepresented groups are modeled almost identically, whereas both comparisons against Caucasians are large. In the unbiased and unbiased~(+) models all three statistics remain below $\sim\!0.025$ regardless of $N$, indicating a latent space that is distributed comparably across the three groups. The $p$-value panel is consistent with this and equally discriminative: the favored-group comparisons of the biased models are overwhelmingly significant, with $p$-values reaching $10^{-40}$ and below, whereas in the unbiased and unbiased~(+) models none of the comparisons is significant. Here, a $p$-value below the threshold indicates a rejection of the null hypothesis ($H_0$), which means that the underlying distributions are statistically distinct. Significance therefore tracks the imposed bias, and the separation between the two regimes is sharp.

\subsubsection{Detection at scale (colored MNIST)}
\label{subsec:res_l2_detection}

We have established that the $KS$ statistic on $p_i$ separates biased from unbiased latent geometries. We turn that statistic into a usable bias \emph{detector} by thresholding its value. Concretely, for each model $\phi$ we compute the three pairwise $KS$ tests between the demographic groups and declare the model biased if at least one of the values $\Gamma$ falls above the chosen threshold. For each model, we compute the clustering probability $p_i$ in the penultimate latent space, separated by color group (red, green, blue), and apply the $KS$ test as in \S\ref{sec:level2}. 

\begin{figure}
\includegraphics[width=\textwidth]{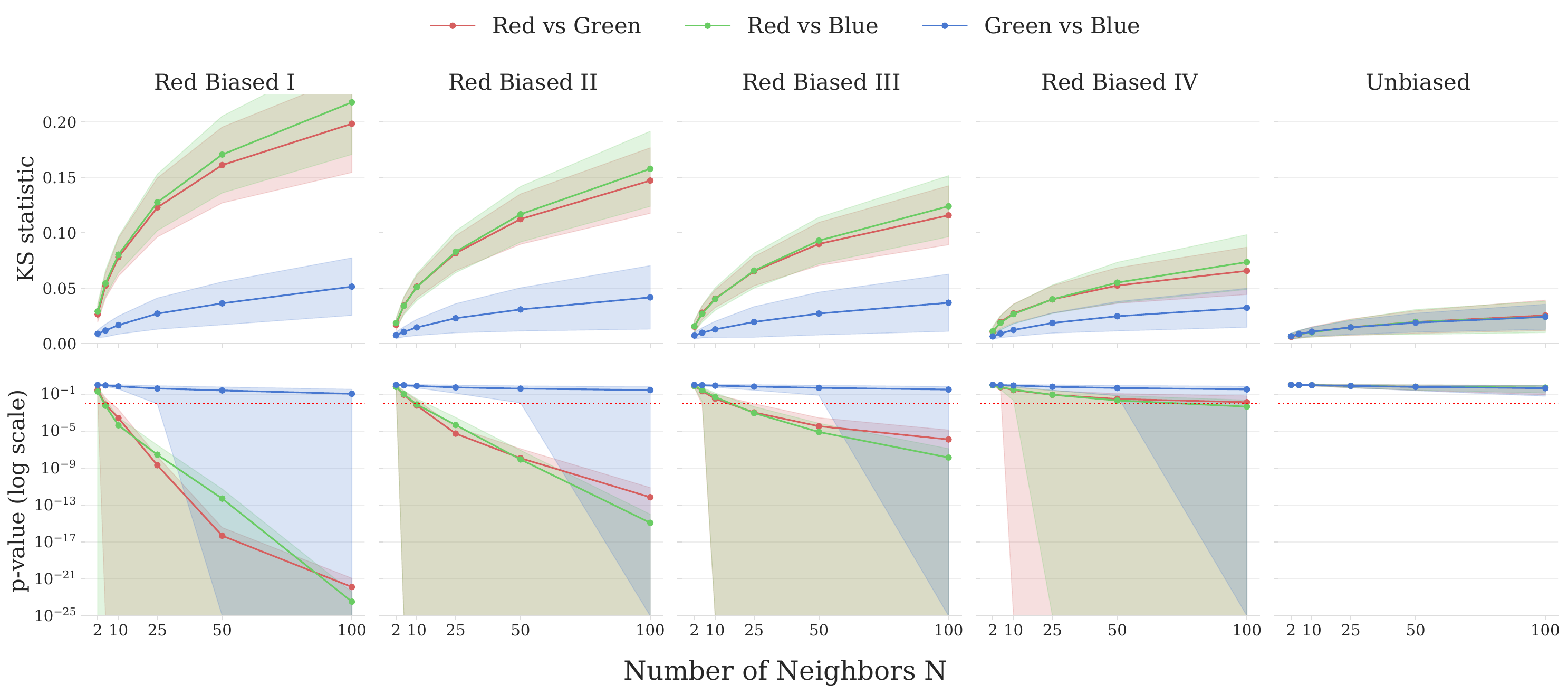}
\caption[]{\textbf{Kolmogorov–Smirnov ($KS$) test statistics and $p$-values evaluated across MNIST models.} Each column shows the average and standard deviation of 100 models for each degree of bias severity, plotting the pairwise statistics and $p$-values of probability distributions between the colored digits against the number of neighbors $N$. The red dotted line marks the rejection threshold for the null hypothesis ($H_0$) of identical distributions with a 99\% confidence level ($\alpha =0.01$); above this line, the null hypothesis cannot be rejected. The $p$-value is in logarithmic scale for better visualization.}
\label{fig:ks_pval_sweep_mnist}
\end{figure}

We used 300 biased models (100 each bias) and 100 unbiased models trained with partition 1 of MNIST. We fix $N=100$ for the headline result. As shown in Figure \ref{fig:ks_pval_sweep_mnist}, the $KS$ decision is stable in the range $10 \le N \le 100$, confirming that the result is not specific to one choice of $N$. We use $\alpha = 0.01$ on the raw $p$-values together with Bonferroni correction over the three pairwise comparisons, giving an effective per-test threshold of $\alpha_{\text{eff}} = 0.01/3 \approx 0.0033$.

We threshold the KS \emph{statistic} rather than its $p$-value for two reasons. The statistic is a bounded effect size that is largely insensitive to sample size, so a single threshold transfers across domains with different test-set sizes (faces, $\sim$6{,}000 per group; MNIST, $\sim$3{,}300 per group), whereas a $p$-value threshold would need recalibration for each. And, as the sweep below shows, the statistic stays clearly discriminative for mild bias, whereas the favored-group $p$-value drifts back toward non-significance as the skew weakens. We fix $N=100$ and set the threshold $\Gamma$ on the largest pairwise statistic so that the false-positive rate in the unbiased models remains at $5\%$, which gives $\Gamma\ge0.06$. In ResNet face models, the same criterion cleanly separates \textit{biased} models (favored-group comparison $\geq 0.12$) from \textit{unbiased} and \textit{unbiased~(+)} ones ($\leq 0.025$).

\begin{figure}[t]
    \centering
    \includegraphics[width=0.55\textwidth]{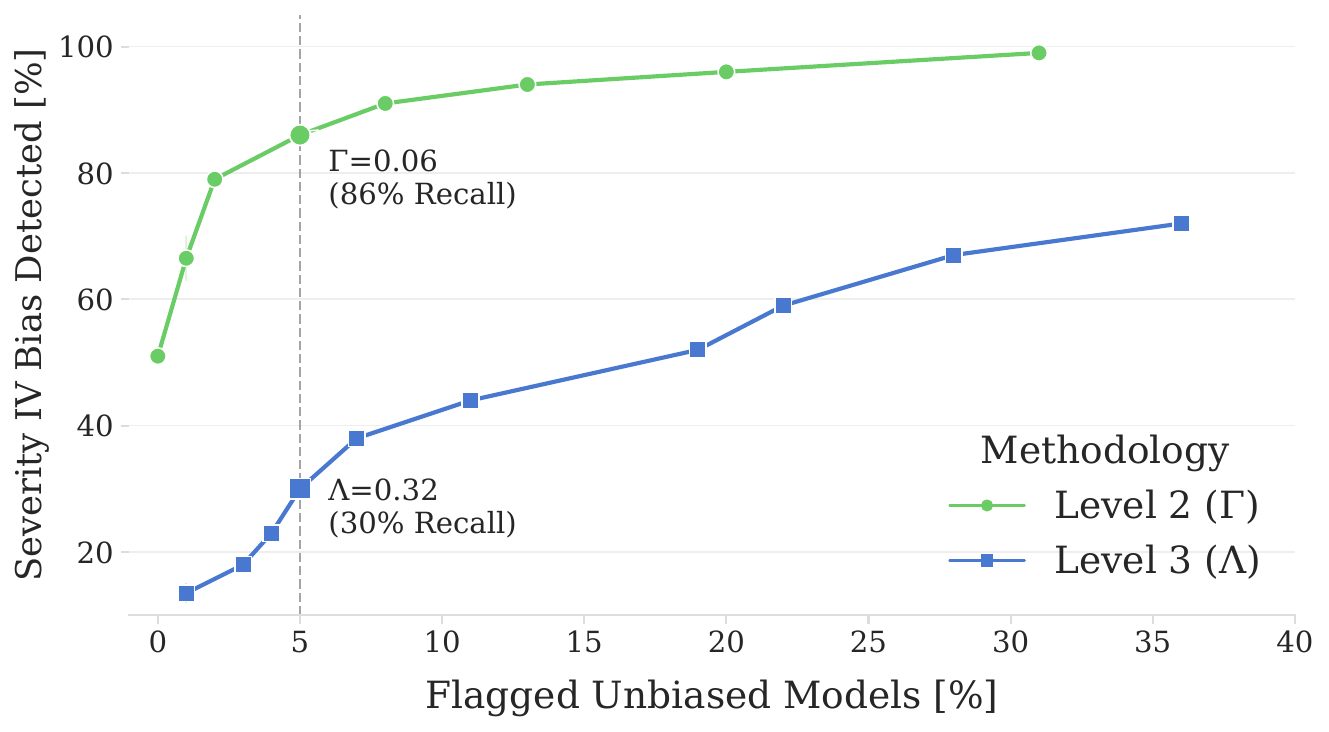}
    \caption[]{\textbf{Detection ROC:} Severity \textbf{IV} Recall vs. Unbiased False Positive Rate. }
    \label{fig:detection_roc}
\end{figure}

In colored MNIST, the favored-color $KS$ statistic at $N=100$ falls steadily from about $0.17$, $0.15$ and $0.12$ to $0.07$ for severities \textbf{I}--\textbf{IV} (Figure~\ref{fig:ks_pval_sweep_mnist}), against about $0.02$ for unbiased models; the comparison between the two minority colors remains small throughout, mirroring the Asian-vs-African/Indian behavior on faces. At the $5\%$-false-positive operating point ($\Gamma\ge0.06$), every model biased at severity levels \textbf{I}--\textbf{III} is detected, and $87\%$ of the mildly biased severity \textbf{IV} models are caught (Table~\ref{tab:mnist_detection}). The choice of operating point matters only at this mild-bias boundary: a more permissive threshold raises severity \textbf{IV} recall to $94\%$ at the cost of a higher false-positive rate, while a stricter one trades recall for specificity (Figure~\ref{fig:detection_roc}). We return to the interpretation of these false positives at the end of \S\ref{subsec:res_l3_detection}.

The statistical test result does not say which group is favored, but that they are different. To see which group is better represented in the latent space, we need to look at the probability distribution. A better representation of the latent space will have the clustering probabilities $p_i$ more concentrated around 1, as in Figure \ref{fig:kde_p_face} (left).

\subsection{Level 3: Activations (ActivationBias)}
\label{subsec:res_l3}

Through this experiment, we observe that the activations of the final convolutional layers serve as indicators of the model's bias. Minority groups, defined by fewer samples during training, exhibit lower activation levels $\lambda$ in the final convolutional layer, whereas groups with a larger number of samples demonstrate higher activation levels $\lambda$. The activation signal differs especially at the end of the network, where the learned patterns are more abstract and task-specific.

\subsubsection{Analysis (ResNet face models)}
\label{subsec:res_l3_analysis}

\begin{figure}[t]
\includegraphics[width=\textwidth]{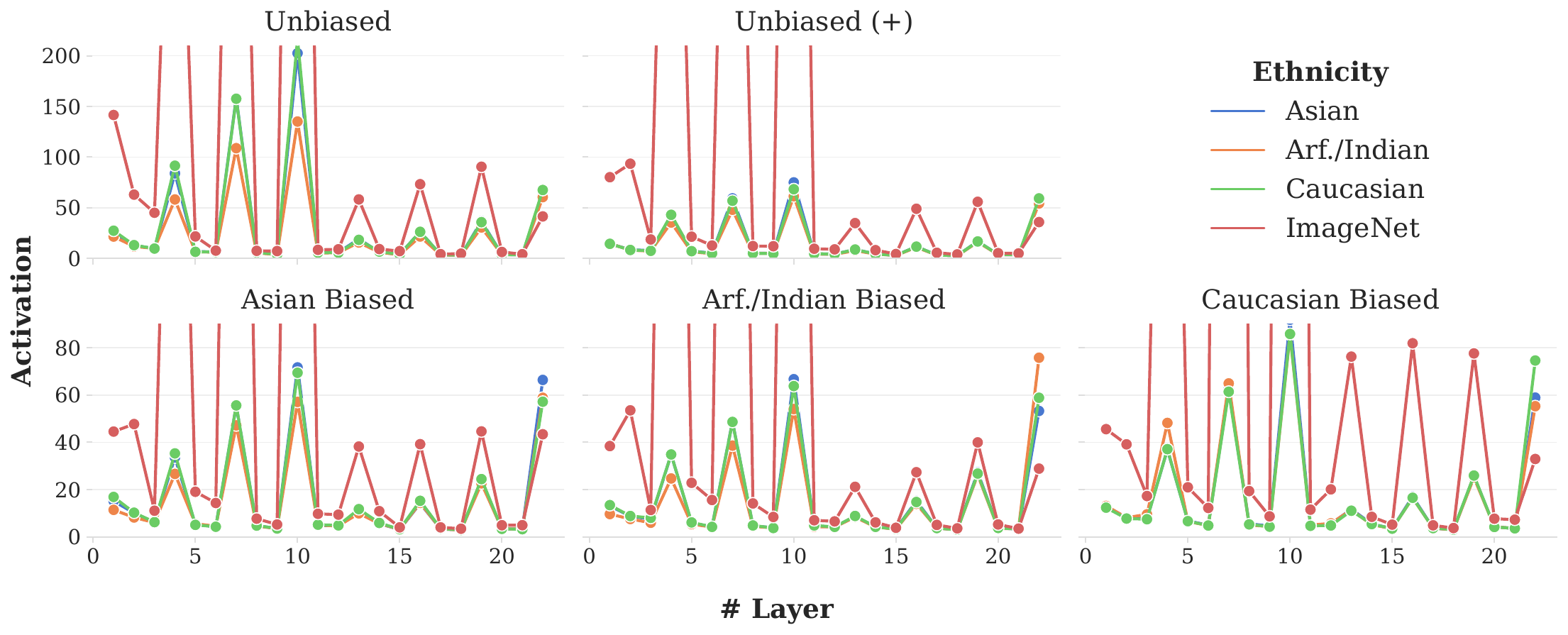}
\caption[]{\textbf{Per-layer activation across the ResNet models.} Group-average activation at every convolutional layer, shown as $10^{\lambda}$, for the three demographic groups (\textit{Asian}, \textit{African/Indian}, \textit{Caucasian}) and, as an out-of-domain control, $1{,}000$ random \textit{ImageNet} images processed identically to the face crops. ($\lambda$ is defined in Equation~\ref{eqn:activation}; the monotonic $10^{\lambda}$ rescaling affects only the display, spreading the wide dynamic range so that the deep-layer differences between groups remain visible.) ImageNet activates the early, generic-feature layers most strongly but decays in the deep, task-specific layers, where the in-domain face groups overtake it; in the \textit{Biased} models the favored group attains the highest activation in the final layer.}
\label{fig:comparativa_biased}
\end{figure}

All activation results on faces use the independent $25\%$ test partition ($6{,}000$ images per ethnicity). Figure \ref{fig:comparativa_biased} reports the group-average activation in every convolutional layer of ResNet models for each demographic group, shown as $10^{\lambda}$ (with $\lambda$ from Eq.~\ref{eqn:activation}) to spread the wide dynamic range and keep the deep-layer differences legible. This monotonic rescaling is only for display; the Activation Bias Indicator and the Mann--Whitney rank statistics are computed on $\lambda$. The per-layer profile shows that the difference in activation between well- and poorly-represented groups is not homogeneous across layers in the \textit{Biased} models, with the final layers of the networks experiencing a particularly prominent trend: the ethnic group prioritized during training, with a higher proportion of data, consistently yields stronger activation in the final layer.

To interpret this trend, the same figure includes an out-of-domain reference: the average activation generated by $1{,}000$ random ImageNet images~\citep{deng2009imagenet}, processed identically to the face crops. ImageNet images activate the early layers more strongly than faces because those layers respond to generic low-level features (edges, colors, and textures) present in any natural image; their activation, however, decays in the deeper layers, which encode the higher-level features specific to the task (gender-relevant facial structure) that ImageNet images lack, and the face images overtake ImageNet precisely there. Because a convolution yields a high response only when the input matches the learned filter and a low one otherwise ~\citep{zeiler2014visualizing}, the activation magnitude measures how well the input fits the patterns the network has actually learned for the task: deep-layer activation reflects the fit of the pattern relevant to the task, rather than a generic property of the image. In light of this, the lower deep-layer activations of the underrepresented groups are not a coincidental correlation, but an interpretable signal that the network has learned their task-relevant patterns less effectively, which in turn may contribute to the performance gaps reported in Table~\ref{tab:compact_models}.

\begin{figure}[t]
\includegraphics[width=\textwidth]{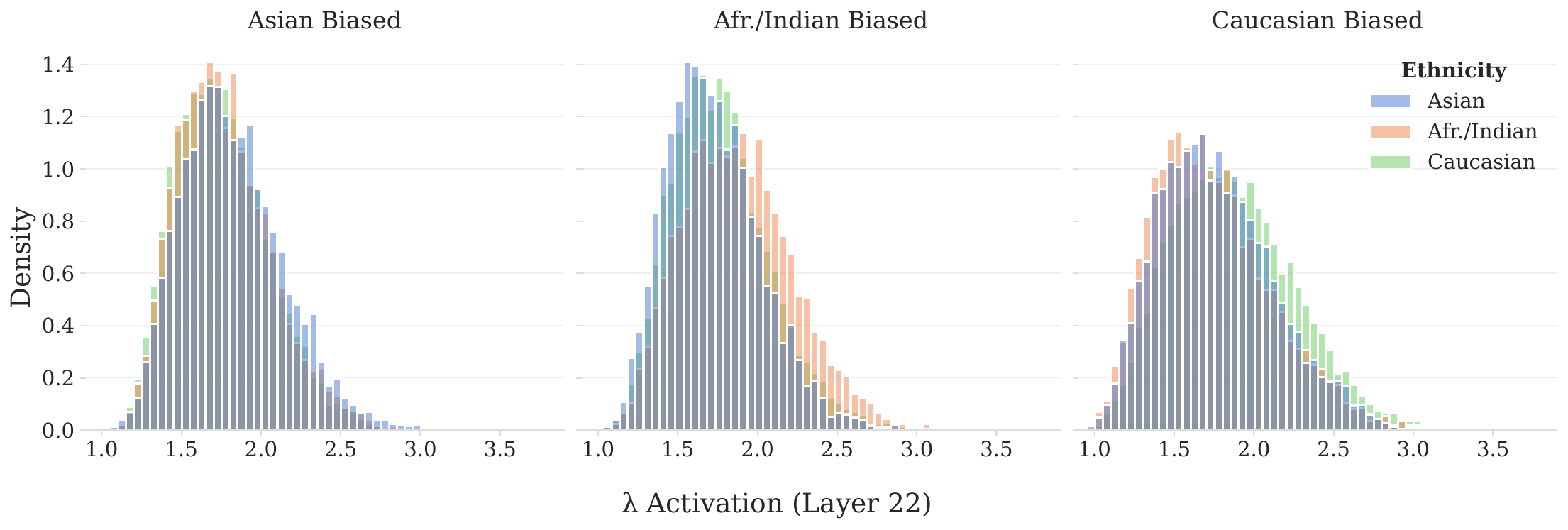}
\caption[]{\textbf{Activation $\lambda$} distribution for the last convolutional layer of the \textit{Biased} ResNet models for the different demographic groups.}
\label{fig:act_dist_face}
\end{figure}

This reading is consistent with the behavior of the early layers. As Table~\ref{tab:compact_models} shows, lower activation in the initial layers does not necessarily translate into lower overall performance: it appears to be compensated by a higher activation in the later layers, which are more closely related to the high-level features of the task $T$ (gender recognition). The bias therefore localizes to the deep, task-specific layers rather than to the early, generic feature extractors. Ultimately, classification draws on the interplay of all activations~\citep{bau2020understanding} rather than on any single layer; what biased models alter is the balance of that interplay in deep layers.

In contrast to \textit{Biased} models, \textit{Unbiased} and \textit{Unbiased~(+)} models (Figure~\ref{fig:comparativa_biased}) exhibit reduced disparities between groups, particularly in the final layers. The \textit{Unbiased~(+)} model, trained with three times more data, shows markedly more uniform activations across demographic groups. These results indicate a coherent relationship between the amount of training data, the performance reported in Table~\ref{tab:compact_models}, and the activations of Figure~\ref{fig:comparativa_biased}: as the training distribution becomes balanced (and larger), the activation disparity in the deep layers shrinks.

Figure~\ref{fig:act_dist_face} examines the activations of the final layer directly: for each biased model, the activation distribution of the favored group is shifted towards higher values relative to the two underrepresented groups, whereas the three distributions coincide for the unbiased models. This is the distributional counterpart of the per-layer trend in Figure~\ref{fig:comparativa_biased}, and is the quantity summarized by the Activation Bias Indicator used for detection.

Because activation is the convolution of the input with the learned filters followed by max pooling---an operation maximized when the input pattern matches the filter and sharpened further by the pooling step---a pattern that occurs frequently in the training data is strongly learned and strongly activated. This is exactly what we observe: the majority group during training attains the highest activation in the final layer, indicating that the network holds filters that closely match the patterns of that group. This direct link between learned filters and activation also motivates the parameter-level analysis of Level~4.

Finally, this signal can be used to detect bias. Bias, however, is intrinsically relative: it describes a relationship between two groups and has no absolute value, so an acceptable level of inequality must be fixed before a model is declared biased---a threshold we make explicit when we turn the activation gap into a detector.

\subsubsection{Severity sweep (colored MNIST)}
\label{subsec:res_l3_sweep}

\begin{figure}[t]
\includegraphics[width=0.4\textwidth]{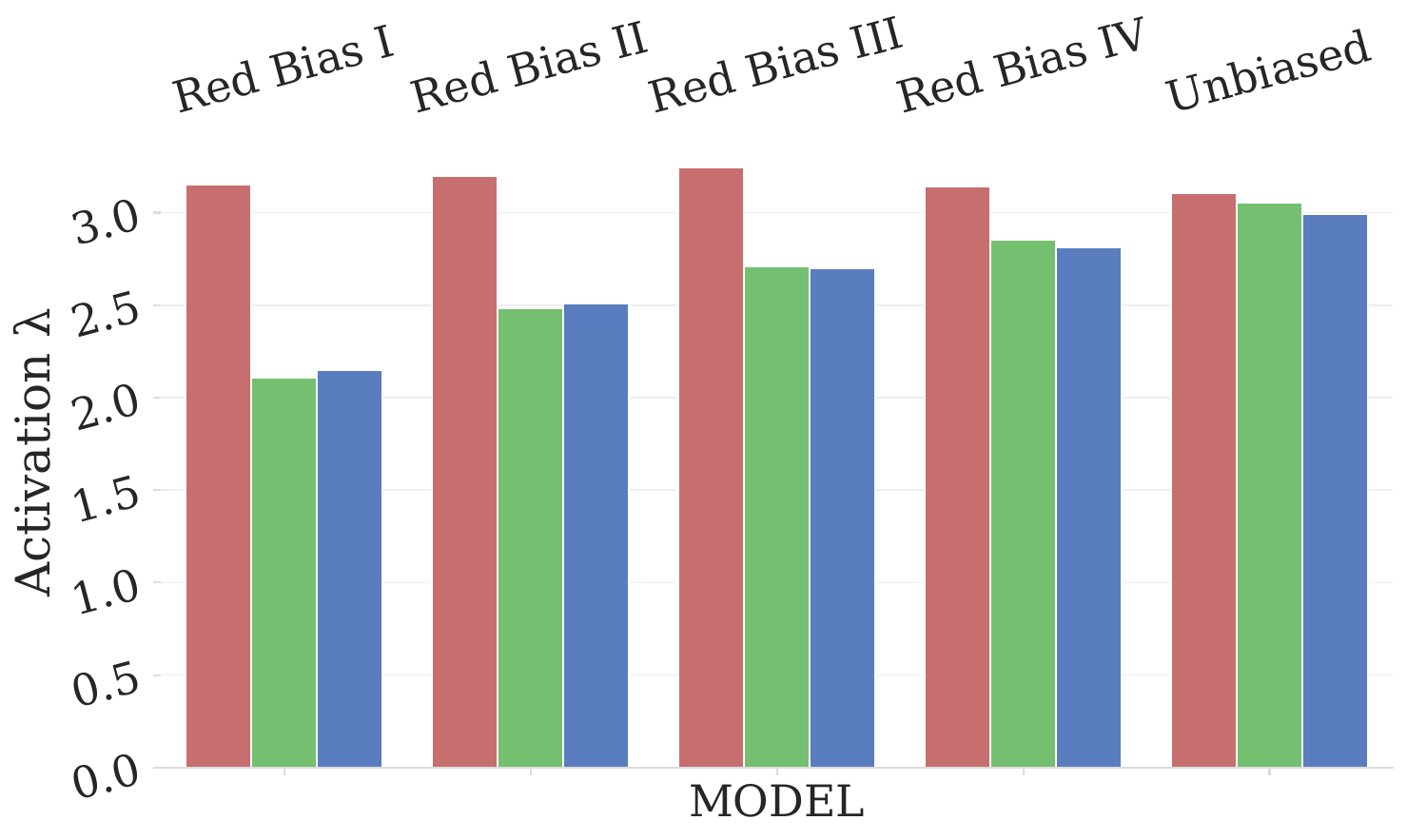}
\includegraphics[width=0.6\textwidth]{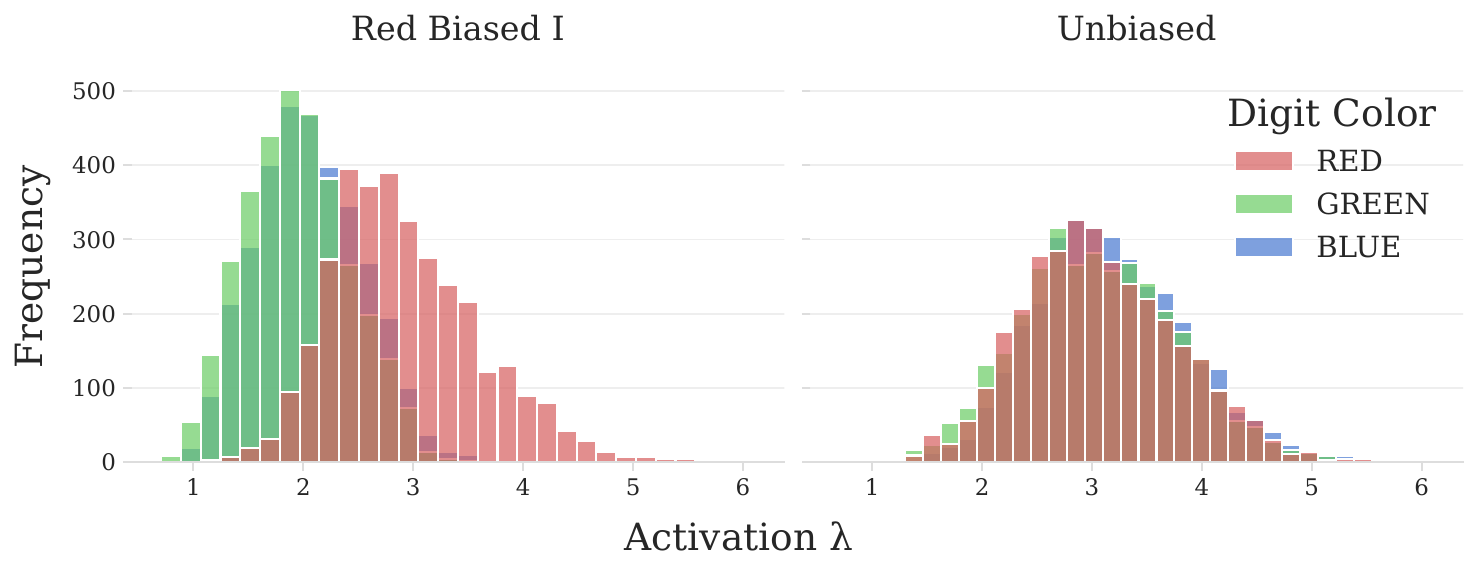}
\caption[]{\textbf{(Left)} Avg. \textbf{Activation $\lambda$} for the last convolutional layer of the Red Biased (severity \textbf{I}-\textbf{IV}) and Unbiased models across digit colors. \textbf{(Right)} \textbf{Activation $\lambda$} distribution for the Red Biased I and Unbiased models for the different digit colors.}
\label{fig:act_sweep_mnist}
\end{figure}
 
Figure~\ref{fig:act_sweep_mnist} (left) reports the group-average final-layer activation by digit color for the red-biased family across the four severities and for the unbiased model (the other two colors  are given in the appendix and are equivalent). The gap between the majority color and the two minority colors is largest at severity~\textbf{I} and shrinks monotonically through severities \textbf{II} and \textbf{III} to severity~\textbf{IV}, and vanishes for the unbiased model, where all three colors activate equally. Figure~\ref{fig:act_sweep_mnist} (right) shows the same effect at the level of distributions: under severity~I the majority color's activations are clearly displaced upward, whereas under the unbiased model the three colors' distributions coincide.

\subsubsection{Detection at scale (colored MNIST)}
\label{subsec:res_l3_detection}
 
For detection we use the Activation Bias Indicator, the Mann--Whitney $U$ comparison of two groups' final-layer activations by its rank-biserial correlation (Eq. \ref{eqn:ABI}), and flag a model biased when its largest pairwise $|\Lambda|$ exceeds a threshold $\rho$. We use this effect size rather than the test's p-value because it is bounded, comparable across domains, and directly measures the magnitude of the activation gap. As at Level~2, $\rho$ trades mild-bias recall against the unbiased false-positive rate, and the full sweep of $\rho$ is given in the appendix: strongly and moderately biased models (severities \textbf{I}--\textbf{III}) are detected at $90\%$ or above across the useful range, while severity-\textbf{IV} recall and the unbiased false-positive rate decline together as $\rho$ grows. Setting $\rho$ to the same $5\%$ false-positive budget gives $\rho=0.32$, at which the detector reaches $99$, $98$, and $90\%$ recall for severities \textbf{I}-\textbf{III} and $30\%$ for severity~\textbf{IV} (Table~\ref{tab:mnist_detection}).

At a matched false positive rate, the two training-free detectors are not equivalent: the latent-space test (Level~2) recovers far more of the mild severity \textbf{IV} bias than the activation test (Level~3)---$86\%$ versus $30\%$ at $5\%$ false positives---indicating that, for weak bias, the structure of the embedding space is a more sensitive signal than the magnitude of the final-layer activations.
 
A final remark concerns the unbiased false positives. Because these detectors measure a property of the learned representation rather than of the training data, the reported false-positive rate is an upper bound on the true error rate: balanced training data does not guarantee a balanced representation. A balanced-data model may still develop a mildly asymmetric internal representation; therefore, some flagged unbiased models may be genuinely, if weakly, biased rather than detector errors, which is why we hold the budget at a conservative $5\%$ rather than at the more permissive thresholds that would capture additional severity \textbf{IV} models.
 
\begin{table}[t]
    \centering
    \caption{\textbf{Bias detection on colored MNIST.} Both training-free detectors operate at a fixed $5\%$ unbiased false-positive rate (Level~2: KS statistic, $N=100$, $\Gamma\ge0.06$; Level~3: rank-biserial correlation, $\Lambda\ge0.32$). Biased rows report recall (fraction of biased models flagged); the unbiased row reports specificity (fraction correctly left unflagged). Values are for the red-biased family; the green- and blue-biased families agree to within a few points. The full per-color and per-threshold results, and the operating-point trade-off (Figure~\ref{fig:detection_roc}), are in the appendix.}
    
    \label{tab:mnist_detection}
    \setlength{\tabcolsep}{12pt}
    \begin{tabular}{@{}lcc@{}}
        \toprule
        \textbf{Configuration} & \textbf{Level 2 (\%)} & \textbf{Level 3 (\%)} \\
        & \emph{$N{=}100,\ \Gamma{=}0.06$} & \emph{$\Lambda{=}0.32$} \\
        \midrule
        Severity I \textit{(99/.5/.5)}    & 100 & 98 \\
        Severity II \textit{(90/5/5)}     & 100 & 94 \\
        Severity III \textit{(80/10/10)}  & 100 & 84 \\
        Severity IV \textit{(60/20/20)}   & 86  & 30 \\
        \midrule
        Unbiased \textit{(33/33/33)}      & 95  & 95 \\
        \bottomrule
    \end{tabular}
\end{table}

\subsection{Detecting Bias in Parameters of the Convolution Filters (Level 4)}
\label{subsec:res_l4}

At this level, a secondary network $\psi$ is trained to classify the bias of a task model only using its convolutional filters, without access to inputs or outputs. The detector therefore requires a population of task models for its own training and evaluation.

\begin{figure*}[t]
    \centering
    \includegraphics[width=\textwidth]{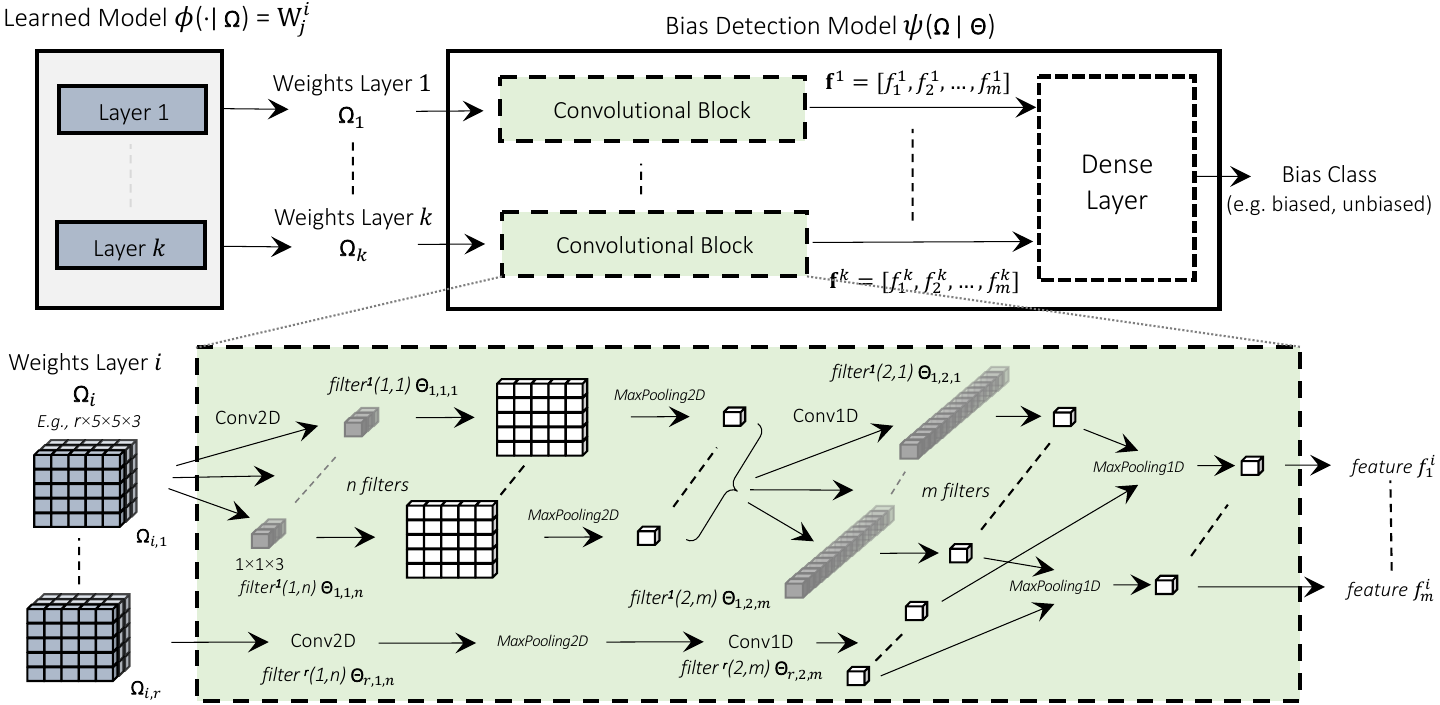}
    \caption{General architecture of a bias detector with the $1\times1$-conv module variant. The architecture depends on the number of layers $k$ of the model $\mathbf{\Omega}$ to be audited. The depth of the module filters depends on the depth of the input weights. Module variant $1\times1$-conv consists of the subsequent layers: $1\times1$ convolution followed by $d\times d$ MaxPooling, then again a one-dimensional convolution with kernel size of $1$ followed by a MaxPooling with pool size equal to the number of input filters.}
    \label{fig:architecture}
\end{figure*}

We devised various bias detection networks $\psi$ that maintain the overall architecture whilst modifying the \textit{Convolutional Block} (Figure \ref{fig:architecture}). The variants outlined here are the outcome of an exploration across a large space of possibilities, and they represent the most informative and noteworthy. The \textit{Convolutional Block} variants analyzed are:

\begin{itemize}
    \item \textbf{MLP}: Flatten $\rightarrow$ Dense($r$)
    \item \textbf{\boldmath$1\times1$ +conv}: Conv2D ($1\times1$) $\rightarrow$ MaxPooling2D ($d\times d$) $\rightarrow$ Conv1D (1) $\rightarrow$ MaxPooling1D ($r$) $\rightarrow$ Flatten.
    \item \textbf{\boldmath$1\times1$ +max}: Conv2D ($1\times1$) $\rightarrow$ MaxPooling3D ($d\times d\times k$) $\rightarrow$ Flatten. 
    \item \textbf{\boldmath$d\times d$ +conv}: Conv2D ($d\times d$) $\rightarrow$ Conv1D (1) $\rightarrow$ MaxPooling1D ($k$) $\rightarrow$ Flatten. 
    \item \textbf{\boldmath$d\times d$ +max}: Conv2D ($d\times d$) $\rightarrow$ MaxPooling1D ($r$) $\rightarrow$ Flatten.
    \item \textbf{\boldmath$1\times1\times1$ +max}: Conv3D ($1\times1\times1$) $\rightarrow$ MaxPooling3D ($d\times d \times c$) $\rightarrow$ Flatten.
    \item \textbf{\boldmath$d\times d \times c$}: Conv3D ($d\times d \times c$) $\rightarrow$ Flatten.
\end{itemize}

\noindent where $d\times d$ represents the dimension of the input filter weights, $c$ denotes the number of input channels, and $r$ refers to the number of input filters. Convolutions are followed by a ReLU activation function, and there is always $0.1$ dropout afterwards (we have seen that it works best among the values: 0.0, 0.1, 0.2 and 0.3).

\paragraph{Face biometrics} In tiny face models, the weight detector recovers the type of demographic bias (three-way: Asian, African/Indian, or Caucasian preferred) from the filters alone, reaching up to $90\%$ accuracy, with the architectures that convolve with a kernel the size of the input weights performing best (d$\times$d-conv $87\%$, d $\times$ d-max $90\%$; Figure~\ref{fig:l4_face_arch} and the ablation in the appendix). This establishes that the demographic bias is encoded in the learned parameters. 

\paragraph{Colored MNIST} In colored MNIST, the detector is trained on the four-way task of identifying the bias condition—red-biased, green-biased, blue-biased, or unbiased (chance = 25\%)---with one detector trained per severity, using the same two architectures that performed best on faces. Table~\ref{tab:l4_mnist_sweep} reports the per-class accuracy. For severities \textbf{I}--\textbf{III}, both architectures are near-perfect on all four classes. At the mildest severity, ~\textbf{IV}, the three biased colors are still recovered well ($93$--$96\%$), but the unbiased class is the one that suffers: its accuracy drops to $78\%$ (d$\times$d-max) and $71\%$ (d $\times$ d-conv) because some mildly biased models are now confused with unbiased ones. This is the weight-level counterpart of the mild-bias confusion seen at Levels~2--3, and carries the same reading: at severity~\textbf{IV} the learned parameters of a biased model are only weakly distinguishable from those of a balanced-data model.

A cross-severity transfer experiment makes the structure of this signature explicit. A detector trained to separate \emph{mildly} biased (severity~\textbf{IV}) models from unbiased ones transfers upward to the strongly biased regime: evaluated on severity-\textbf{I} biased and unbiased models it reaches $97\%$ accuracy. The reverse completely fails: a detector trained in severity~\textbf{I} becomes completely blind to mild bias when tested on severity~\textbf{IV}, classifying every model as unbiased (100\% detection of unbiased models, 0\% for biased models). The weight signature of strong bias is thus an amplified version of that of mild bias: a detector tuned to the subtle signature also recognizes the obvious one, whereas a detector tuned to the obvious signature collapses to predicting ``unbiased'' once the signature is faint. In practice, this argues for training a weight-based detector on the mildest bias one intends to catch.

This result demonstrates the existence of identifiable patterns associated with bias within the weights of a neural network. The fact that a neural network $\psi$ can accurately classify the weights of another neural network $\phi$ based on the type of bias implies that there is a discernible pattern encoded in the weights of network $\phi$ related to bias.

\begin{table}[t]
\centering
\caption{\textbf{Colored-MNIST \textit{Level~4}: four-way weight detector.} Per-class accuracy (\%) of the classifier that identifies which color, if any, was over-represented (red-, green-, blue-biased, or unbiased; chance $=25\%$), with one detector trained per severity, for the two best architectures. Strong and moderate bias (\textbf{I}--\textbf{III}) is recovered almost perfectly; at severity~\textbf{IV} the biased classes remain well identified while the unbiased class is increasingly confused with mild bias.}
\label{tab:l4_mnist_sweep}
\begin{tabular}{@{}lcccc@{}}
\toprule
\textbf{Severity} & \textbf{Red Biased} & \textbf{Green Biased} & \textbf{Blue Biased} & \textbf{Unbiased} \\
\midrule
\multicolumn{5}{@{}l}{\textit{d$\times$d-max}} \\
\quad I \textit{(99/.5/.5)}   & 100.00 & 100.00 & 100.00 & 100.00 \\
\quad II \textit{(90/5/5)}    & 99.85  & 99.88  & 99.94  & 99.94  \\
\quad III \textit{(80/10/10)} & 99.20  & 99.85  & 99.00  & 98.45  \\
\quad IV \textit{(60/20/20)}  & 94.25  & 97.80  & 93.45  & 88.05  \\
\midrule
\multicolumn{5}{@{}l}{\textit{d$\times$d-conv}} \\
\quad I \textit{(99/.5/.5)}   & 99.95  & 99.90  & 99.90  & 99.90  \\
\quad II \textit{(90/5/5)}    & 99.84  & 99.54  & 99.83  & 98.37  \\
\quad III \textit{(80/10/10)} & 96.65  & 99.15  & 98.30  & 96.95  \\
\quad IV \textit{(60/20/20)}  & 91.40  & 96.45  & 95.00  & 83.50  \\
\bottomrule
\end{tabular}
\end{table}

\section{Conclusions}
\label{sec:conclusion}

The ability to detect and quantify biases in neural networks has significant implications for the field of computer vision and AI ethics. It enables developers to create fairer and more inclusive AI systems by identifying and mitigating biases that could lead to discriminatory outcomes. This study presents a robust methodology for the detection and quantification of biases in Convolutional Neural Networks (CNNs). The methodology is then applied to the context of facial biometrics in the gender classification task and Colored-MNIST digits. It is based on the examination of embeddings, layer activations, and convolutional filters, meaning that it focuses on the internal workings of neural networks rather than just their output. 

Traditional methods for bias detection, like dataset audits and performance evaluation, while useful, have limitations in terms of scope and data requirements; they often rely on performance metrics across groups, requiring extensive labeled datasets. In contrast, our approach focuses on the internal representations and learned parameters of neural networks, offering two advantages: i) a more granular insight into where and how biases manifest within the network, enabling targeted interventions, and ii) the ability to detect biases with fewer data samples, making it more efficient and practical, particularly in scenarios where large balanced datasets are not available. 

First, our exploration of the learned latent space prior to the classification layer provides deeper insights into how biases affect the distribution of representations. We demonstrate that biased models exhibit a poor distribution of underrepresented groups within the latent space, which directly impacts classification accuracy. 

Second, examining activations of the network layers further illustrates how bias manifests within the model. We observe that groups underrepresented in the training data consistently exhibit lower activation levels in the final convolutional layers compared to majority groups.

Finally, our analysis of convolutional filters reveals that CNNs encode demographic biases within their filters. This encoding is apparent from the high accuracy of our bias detection model, which can classify biases looking at the convolutional filters alone. Although this method offers independence from the input data, it requires training thousands of models and is less transferable to other architectures or tasks.

The methodologies of activations and latent space are applicable to various neural network architectures and can be extended to different types of bias. Building on our detection methodology, which facilitates the understanding and debugging of biases, it is possible to devise strategies to mitigate identified biases. These strategies could potentially be incorporated into the training loss function and the evaluation phases of neural network development.

Across the three levels, bias leaves a consistent and detectable trace in the internals of the network: a less uniformly distributed latent space (Level~2), lower activation of the underrepresented groups in the final convolutional layers (Level~3), and a recoverable signature in the convolutional filters themselves (Level~4). The colored-MNIST severity sweep ties the levels together: at every level, the disparity and with it the detector's discriminative power, contracts smoothly as the training distribution moves from skewed toward balanced, vanishing in the unbiased models.

Table \ref{tab:method_comparison} summarizes key qualitative characteristics of  traditional methods and those proposed here. Our methods complement existing approaches by filling gaps in bias detection where traditional methods may fall short. Although our methods are not universally superior, they contribute to the field by providing alternative tools for bias detection, particularly in scenarios with limited data or where deeper insights into network behavior are required. The table highlights these comparative strengths and weaknesses, situating our work within the broader landscape of bias detection methods.

\section{Limitations and Future Work}

Several limitations should be noted. (1) The analysis focuses on convolutional architectures; extending the framework to transformer-based architectures is left for future work. (2) The proposed indicators depend on methodological choices (neighborhood size N, significance level, and detection thresholds); we characterize their sensitivity (Section 5.3 and Appendix B), but a single setting need not be optimal for every model or task. (3) Level-2 and Level-3 signals are statistical properties of internal representations and should be read as indicators correlated with, rather than proof of, discriminatory output behavior. (4) The Level-4 weight-based detector requires training a population of task models, which is computationally demanding; we mitigate this with small, inexpensive models and cross-severity transfer, but the cost still limits applicability to very large models.

In future work, we will exploit recent advances in AI interpretability to evaluate biases \citep{serna2025latent} and adapt the bias analysis-detection methodologies developed here to LLMs \citep{pena25llms} and multimodal setups combining language and vision models \citep{2020_ICMI_FairDemo_Pena}.

\begin{table}[]
\centering
\caption{Comparison of bias detection methods: traditional and proposed here.}
\resizebox{1\textwidth}{!}{ 
\begin{tabular}{cclll}
\toprule
\textbf{\begin{tabular}[c]{@{}c@{}}Method\end{tabular}}  &
  \multicolumn{1}{c}{\textbf{Level}} &
  \multicolumn{1}{c}{\textbf{Pros}} &
  \multicolumn{1}{c}{\textbf{Cons}} &
  \multicolumn{1}{c}{\textbf{Applicability}} \\ \midrule
  
\textit{\begin{tabular}[c]{@{}c@{}}Dataset \\ Audits\end{tabular}}  &
  \begin{tabular}[c]{@{}l@{}} 1 \end{tabular} &
  \begin{tabular}[c]{@{}l@{}}- Insights into data distribution\end{tabular} &
  \begin{tabular}[c]{@{}l@{}}- Confined to dataset bias\end{tabular} &
  \begin{tabular}[c]{@{}l@{}}Effective for initial \\ dataset bias detection\end{tabular} \\ \midrule

\textit{\begin{tabular}[c]{@{}c@{}}Performance \\ Evaluation \end{tabular}}  &
  \begin{tabular}[c]{@{}l@{}} 1 \end{tabular} &
  \begin{tabular}[c]{@{}l@{}}- Direct evaluation of model fairness\end{tabular} &
  \begin{tabular}[c]{@{}l@{}}- Biases can be compensated \\ and remain undetected \end{tabular} &
  \begin{tabular}[c]{@{}l@{}}Generalizable to any \\ task and model\end{tabular} \\ \midrule \midrule 
  
\begin{tabular}[c]{@{}c@{}}\textbf{SpaceBias}\end{tabular}&
  \begin{tabular}[c]{@{}l@{}} 2 \end{tabular} &
  \begin{tabular}[c]{@{}l@{}}- Captures internal representations\\ - Uses statistical validation\\ - Training free\end{tabular} &
  \begin{tabular}[c]{@{}l@{}}- Needs calibration \end{tabular} &
  \begin{tabular}[c]{@{}l@{}}Ideal for analyzing \\ any latent space \end{tabular} \\ \midrule
  
\begin{tabular}[c]{@{}c@{}}\textbf{ActivationBias}\end{tabular}&
  \begin{tabular}[c]{@{}l@{}} 3 \end{tabular} &
  \begin{tabular}[c]{@{}l@{}}- Insights into model behavior\\ - Uses statistical validation\\ - Training free\end{tabular} &
  \begin{tabular}[c]{@{}l@{}}- Needs calibration\end{tabular} &
  \begin{tabular}[c]{@{}l@{}}Ideal for analyzing \\ any layer activation\end{tabular} \\ \midrule

\begin{tabular}[c]{@{}c@{}}\textbf{WeightBias}\end{tabular}&
  \begin{tabular}[c]{@{}l@{}} 4 \end{tabular} &
  \begin{tabular}[c]{@{}l@{}}- Independence from the input\end{tabular} &
  \begin{tabular}[c]{@{}l@{}}- Requires training models\\ - Not transferable\end{tabular} &
  \begin{tabular}[c]{@{}l@{}}Reduced\end{tabular} \\ \bottomrule
\end{tabular}
}
\label{tab:method_comparison}
\end{table}

\bibliographystyle{elsarticle-num-names} 
\bibliography{references}

\clearpage
\appendix

\section{Model Architectures}
\label{appx:architecture}

\begin{table}[h!]
\centering
\renewcommand{\arraystretch}{1.2}
\begin{tabular}{@{}llllc@{}}
\toprule
\textbf{Layer} & \textbf{Type} & \textbf{Configuration} & \textbf{Output Features} & \textbf{Activation} \\
\midrule
\textbf{Input} & Input & Shape: $28 \times 28 \times 3$ & - & - \\
\textbf{1} & Conv2D & Kernel: $5 \times 5$ & 16 & ReLU \\
\textbf{2} & MaxPooling2D & Pool: $2 \times 2$ & 16 & - \\
\textbf{3} & Conv2D & Kernel: $3 \times 3$ & 32 & ReLU \\
\textbf{4} & MaxPooling2D & Pool: $2 \times 2$ & 32 & - \\
\textbf{5} & Conv2D & Kernel: $3 \times 3$ & 64 & ReLU \\
\textbf{6} & MaxPooling2D & Pool: $2 \times 2$ & 64 & - \\
\textbf{7} & Flatten & - & - & - \\
\textbf{8} & Dense & Fully Connected & 64 & ReLU \\
\textbf{9} & Dropout & Rate: 0.3 & - & - \\
\textbf{10} & Dense & Fully Connected & 10 & Softmax \\
\bottomrule
\end{tabular}
\caption{Architecture of the MNIST Classification Model.}
\end{table}

This section details the architectures of the three convolutional neural networks (CNNs) utilized in this study: a small CNN for MNIST digit classification, a lightweight CNN for Gender Classification, and a custom Residual Network (ResNet) for robust Gender Classification.

\subsection*{Colored-MNIST Classification Model}
\label{appx:mnist_arch}

The MNIST model employs a standard Sequential CNN architecture. Notably, while standard MNIST images are grayscale, this model is configured to accept three-channel inputs ($28 \times 28 \times 3$), an adaptation for RGB-formatted digit images. The network sequentially applies convolutional layers with increasing filter counts to extract hierarchical features, followed by aggressive spatial downsampling via max pooling. A dropout layer is introduced before the final classification head to mitigate overfitting. It contains a total of 29,162 parameters.

\begin{table}[h!]
\centering
\renewcommand{\arraystretch}{1.2}
\begin{tabular}{@{}llllc@{}}
\toprule
\textbf{Layer} & \textbf{Type} & \textbf{Configuration} & \textbf{Output Features} & \textbf{Activation} \\
\midrule
\textbf{Input} & Input & Shape: $120 \times 120 \times 3$ & - & - \\
\textbf{1} & Conv2D & Kernel: $5 \times 5$ & 24 & ReLU \\
\textbf{2} & MaxPooling2D & Pool: $2 \times 2$ & 24 & - \\
\textbf{3} & Conv2D & Kernel: $3 \times 3$ & 48 & ReLU \\
\textbf{4} & MaxPooling2D & Pool: $2 \times 2$ & 48 & - \\
\textbf{5} & Conv2D & Kernel: $3 \times 3$ & 48 & ReLU \\
\textbf{6} & MaxPooling2D & Pool: $2 \times 2$ & 48 & - \\
\textbf{7} & Conv2D & Kernel: $3 \times 3$ & 64 & ReLU \\
\textbf{8} & MaxPooling2D & Pool: $2 \times 2$ & 64 & - \\
\textbf{9} & Conv2D & Kernel: $3 \times 3$ & 64 & ReLU \\
\textbf{10} & MaxPooling2D & Pool: $2 \times 2$ & 64 & - \\
\textbf{11} & Flatten & - & - & - \\
\textbf{12} & Dense & Fully Connected & 128 & ReLU \\
\textbf{13} & Dense & Fully Connected & 2 & Softmax \\
\bottomrule
\end{tabular}
\caption{Architecture of the Tiny Gender Classification Model.}
\end{table}

\subsection*{Gender Classification Face Biometrics (Tiny)}

The ``Tiny'' gender classification model is a lightweight, strictly sequential CNN designed for processing $120 \times 120 \times 3$ facial images. It features a deeper architecture than the MNIST model, using five convolutional blocks. The initial layer uses a larger $5 \times 5$ receptive field to capture broad spatial features, while subsequent layers rely on standard $3 \times 3$ kernels to extract finer biometric details. The spatial dimensions are heavily compressed through five consecutive max-pooling operations prior to classification. It comprises 106,242 parameters.

\subsection*{Gender Classification Face Biometrics (ResNet)}

The ResNet-based gender classification model is a truncated custom variant of the standard Residual Network architecture, utilizing ``bottleneck'' blocks. It contains \textbf{371,682 total parameters} (366,626 trainable) and operates on $120 \times 120 \times 3$ input images. The network relies heavily on Batch Normalization and skip connections to optimize gradient flow and allow deeper feature extraction without degradation. 

The architecture is divided into three distinct macroscopic stages:

\begin{itemize}
    \item \textbf{Stem:} The network begins with a $7 \times 7$ Convolution (32 filters) with zero padding, followed by Batch Normalization, ReLU activation, and MaxPooling2D operation. This reduces the initial spatial resolution to $30 \times 30$.
    \item \textbf{Residual Stage 1:} Contains three bottleneck blocks that operate at $30 \times 30$ spatial resolution. 
    \begin{itemize}
        \item The first block is a convolutional block with a $1 \times 1$ skip connection (expanding the channel depth from 32 to 128). 
        \item The subsequent two are identity blocks. Each bottleneck uses a progression: $1 \times 1$ (32 filters) $\rightarrow$ $3 \times 3$ (32 filters) $\rightarrow$ $1 \times 1$ (128 filters).
    \end{itemize}
    \item \textbf{Residual Stage 2:} Contains four bottleneck blocks. 
    \begin{itemize}
        \item The first is a convolutional block for downsampling that reduces the spatial dimensions to $15 \times 15$ and expands the channel dimension to 256. 
        \item The remaining three are identity blocks. The bottleneck structure here follows a progression: $1 \times 1$ (64 filters) $\rightarrow$ $3 \times 3$ (64 filters) $\rightarrow$ $1 \times 1$ (256 filters) progression.
    \end{itemize}
    \item \textbf{Classification Head:} A Global Average Pooling2D layer maps the $15 \times 15 \times 256$ tensor to a flat 256-dimensional vector, feeding into a final Dense layer of 2 units (softmax) for binary classification.
\end{itemize}

\newpage

\noindent \textbf{Summary of ResNet Macro-Architecture:}

\begin{table}[h!]
\centering
\renewcommand{\arraystretch}{1.2}
\begin{tabular}{@{}llll@{}}
\toprule
\textbf{Stage} & \textbf{Operations} & \textbf{Output Shape} & \textbf{Channels} \\
\midrule
\textbf{Input} & Raw Image & $120 \times 120$ & 3 \\
\textbf{Stem} & $7 \times 7$ Conv2D, BN, ReLU, MaxPool & $30 \times 30$ & 32 \\
\textbf{Stage 1} & $1 \times$ Conv Block + $2 \times$ Identity Blocks & $30 \times 30$ & 128 \\
\textbf{Stage 2} & $1 \times$ Conv Block + $3 \times$ Identity Blocks & $15 \times 15$ & 256 \\
\textbf{Head} & Global Average Pooling2D, Dense(2) & $1 \times 1$ & 2 \\
\bottomrule
\end{tabular}
\caption{Macro-Architecture of the ResNet Gender Classification Model.}
\end{table}

\subsection{Implementation Details and Reproducibility}

\paragraph{Training configuration} All task models (ResNet face, tiny face, and colored-MNIST) were trained from scratch with Glorot-uniform initialization using the Adam optimizer (learning rate 0.01, [schedule, e.g., constant / reduce-on-plateau, factor [..], patience [..]]), batch size 32, for 10-40 epochs (tiny face and colored-MNIST - ResNet face), minimizing the categorical cross-entropy loss.

\paragraph{Preprocessing} Faces were detected and aligned using RetinaFace\cite{deng2020retina}, resized to 120×120×3, and normalized; no augmentation was used. Colored-MNIST images were converted to RGB at 28×28×3 by coloring the background (red, green, or blue) while leaving the digit strokes white, with pixel values normalized.

\section{Robustness to Evaluation Sample Size (Face Biometrics)}
\label{appx:few_samples}

The training-free detectors are computed from a finite set of evaluation images, so it is worth asking how much data they require. On the five ResNet models we recompute the Level-2 and Level-3 signals from random subsets of the test set---$100$ and $1{,}000$ images per ethnicity ($50$ and $500$ per gender)---and repeat each setting over $100$ independent random draws to obtain the distribution of outcomes.

\begin{figure}[t]
    \centering
    \includegraphics[width=\columnwidth]{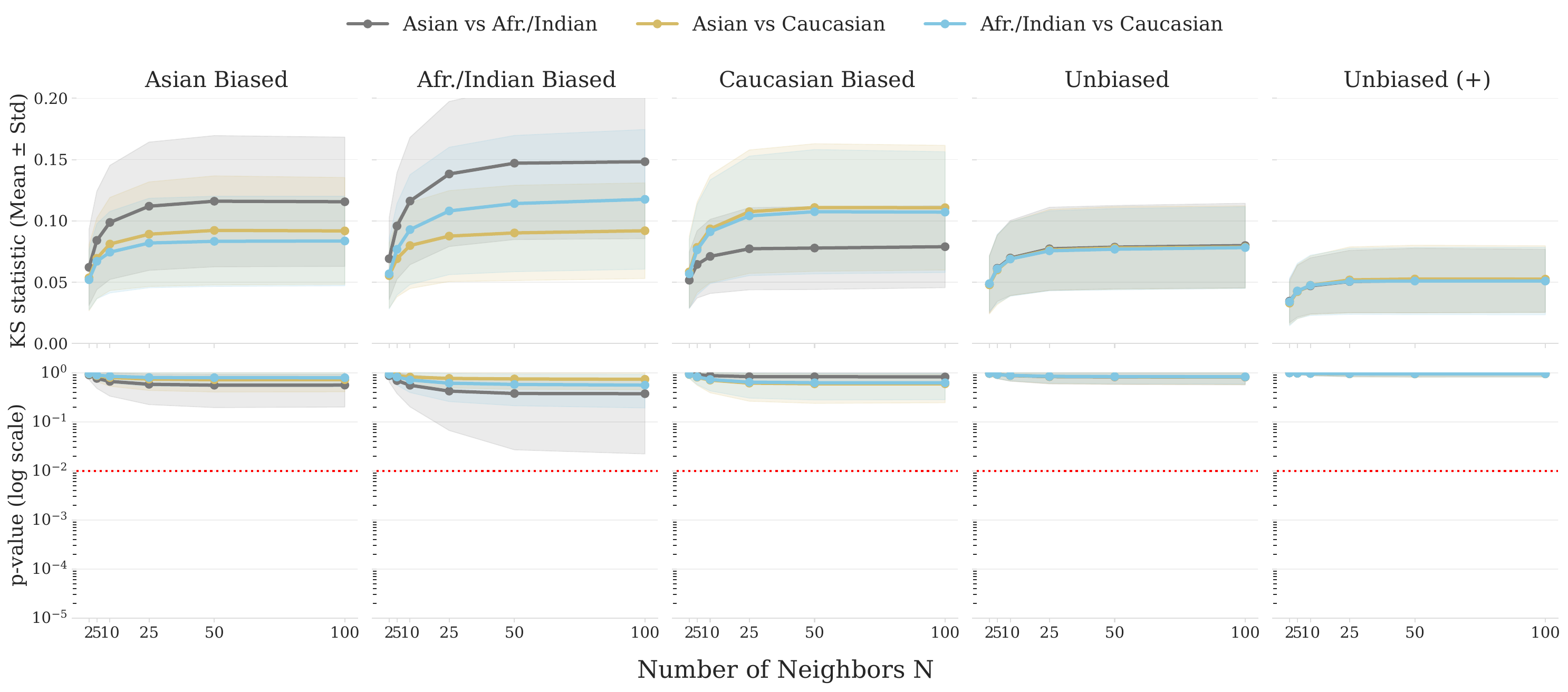}
    \caption{\textbf{Kolmogorov–Smirnov ($KS$) test statistics and $p$-values evaluated across ResNet models.} Each plots shows the average and standard deviation of 100 simulations with 100 samples per simulation. The $p$-value lacks the power to reject at this size.}
    
    \label{fig:ks_sampled_100}
\end{figure}

\begin{figure}[th!]
    \centering
    \includegraphics[width=\columnwidth]{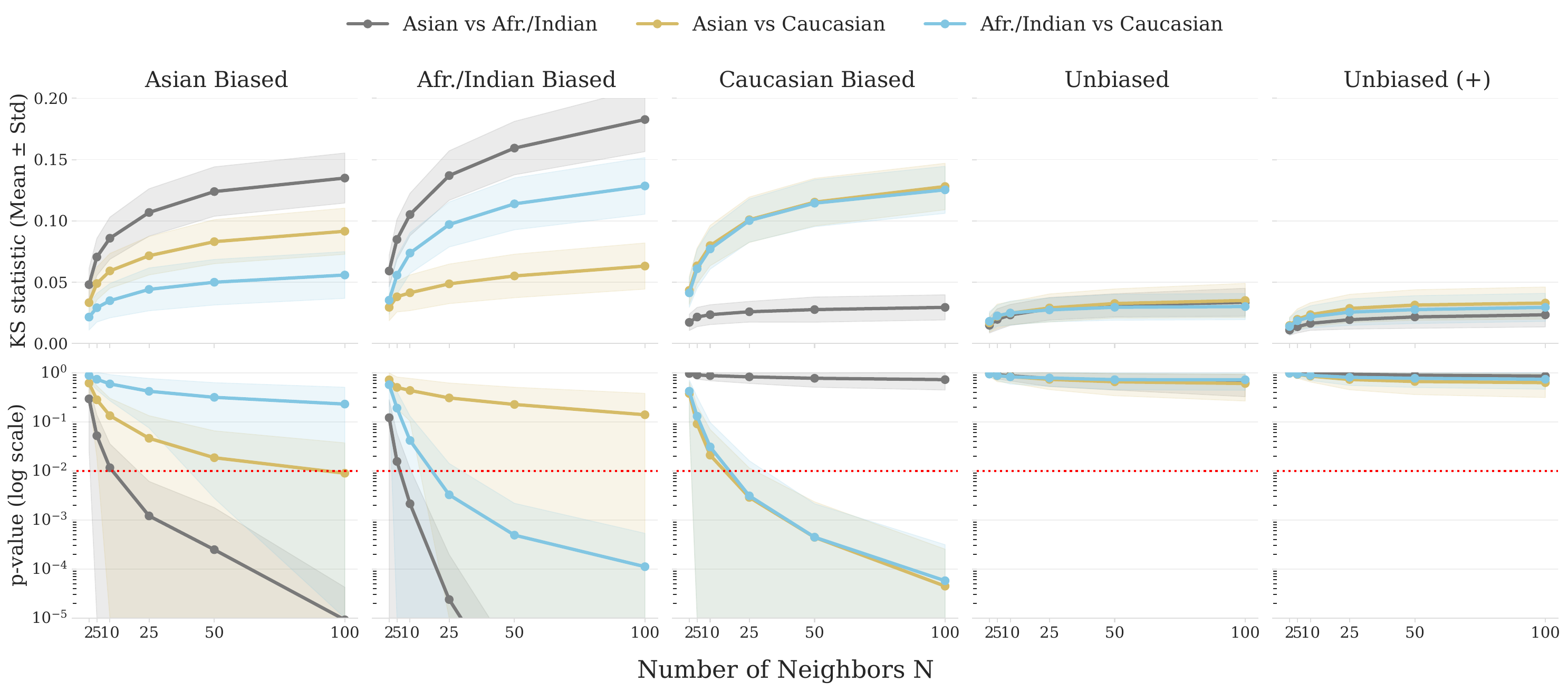}
    \caption{\textbf{Kolmogorov–Smirnov ($KS$) test statistics and $p$-values evaluated across ResNet models.} Each plots shows the average and standard deviation of 100 simulations with 1,000 samples per simulation.}
    \label{fig:ks_sampled_1,000}
\end{figure}
\begin{figure}[h!]
    \centering
    \includegraphics[width=0.49\columnwidth]{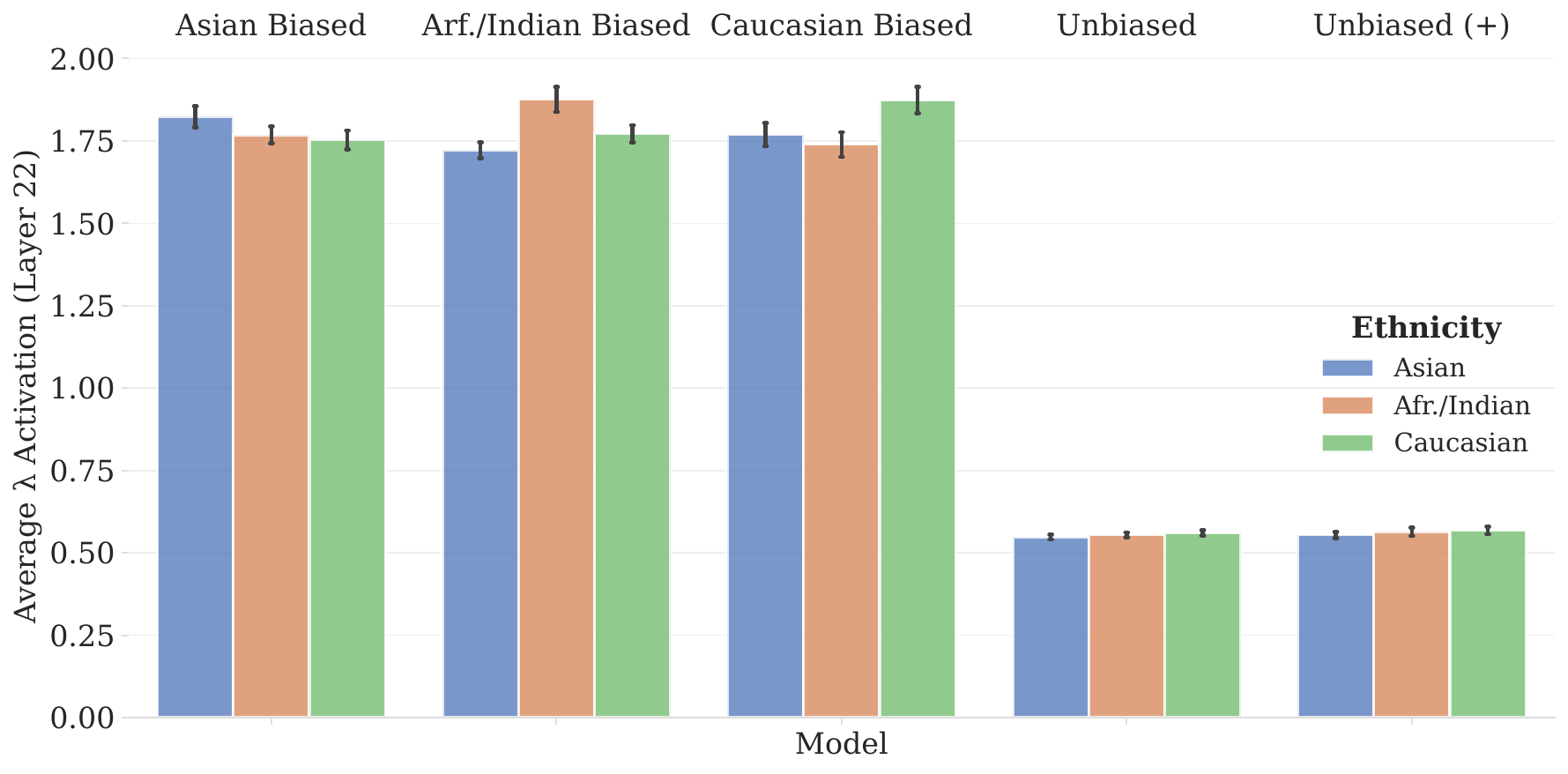}
    \includegraphics[width=0.49\columnwidth]{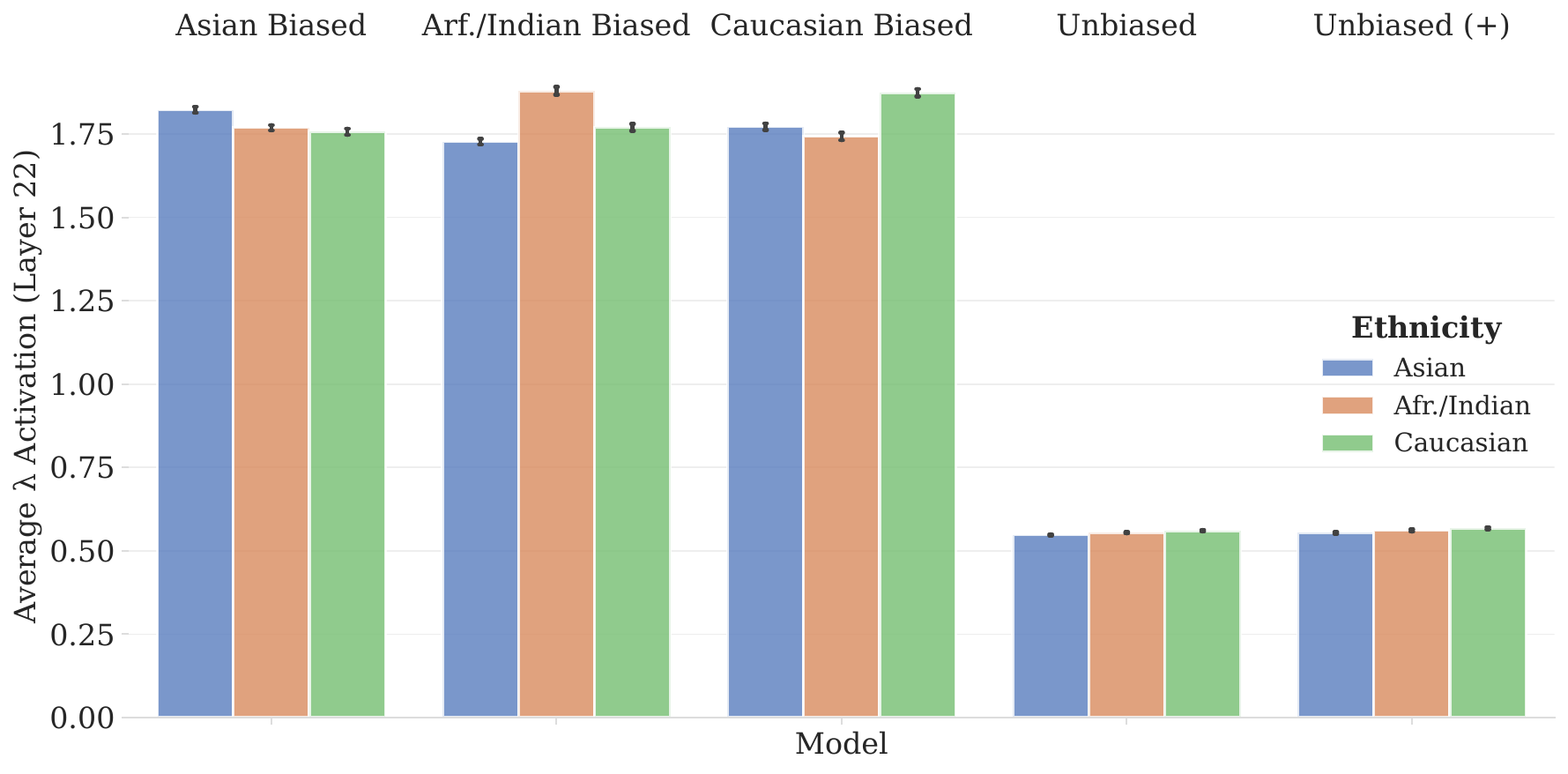}
    \caption{\textbf{Avg. Activation $\lambda$} for 100 sample size (left) and 1,000 sample size (right).}
    \label{fig:act_bars}
\end{figure}

\paragraph{Latent space (Level 2)} Figure~\ref{fig:ks_sampled_100} shows the Level-2 analysis on the $100$-image draws (mean $\pm$ standard deviation over the draws). At this size the Kolmogorov--Smirnov $p$-value has no power: for every model, including the biased ones, the $p$-values of all three pairwise comparisons remain above the $0.01$ significance line at every neighborhood size, so the test does not reject. The $KS$ statistic still orders the models, but the unbiased baseline increases from $\sim\!0.02$ on the full test set to $\sim\!0.05$--$0.08$, eroding the margin on which the threshold at-scale ($\Gamma=0.06$) relies. Increasing the subset to $1{,}000$ images contracts the spread and restores the separation (Figure~\ref{fig:ks_sampled_1,000}).
 
\paragraph{Activations (Level 3)} Figures~\ref{fig:act_bars} (a) and (b) show the per-group average final-layer activation $\lambda^{[l]}$ for the two sizes of subsets. The group means are essentially unchanged between them -- the favored group is the highest in every biased model, and the three groups coincide for the unbiased and unbiased~(+) models---while the standard deviation across draws contracts by roughly $\sqrt{10}$ from the $100$- to the $1{,}000$-image subsets, the $1/\sqrt{n}$ concentration expected of a sample mean. The same contraction holds for the $KS$ statistic, so neither signal is intrinsically more stable than the other; both simply sharpen with more data.
 
\paragraph{Automatic detection} Table~\ref{tab:few_samples_detection} reports, for each subset size, the fraction of the $100$ draws in which each model is flagged biased by the Activation Bias Indicator at the $\Lambda \geq 0.2$ and $\Lambda \geq 0.1$ operating points. With $100$ images the more strongly biased models are detected in a majority of draws ($60$--$80\%$) but the weakly biased Asian model is not ($24\%$), and the two unbiased models flag at a similar low rate ($10$--$14\%$). With $1{,}000$ images, every biased model is detected ($83$--$100\%$), and the contrast between the two unbiased models emerges: the standard unbiased model is flagged in $52\%$ of draws while the unbiased~(+) model is flagged in only $2\%$. Because the estimate is tight at this size, this is not sampling noise but evidence that the standard unbiased model retains a small, real representational asymmetry that the additional balanced data of the unbiased~(+) model removes.
 
\begin{table}[t]
\centering
\caption{\textbf{Robustness to evaluation sample size (ResNet face models).} Flag rate---the percentage of $100$ random draws in which each model is flagged biased by the Activation Bias Indicator at $\Lambda \geq 0.2$ and $\Lambda \geq 0.1$---for two per-ethnicity subset sizes. Strong bias is detected at both sizes; the weakly biased Asian model and the residual asymmetry of the standard unbiased model become apparent only at the larger size.}
\label{tab:few_samples_detection}
\setlength{\tabcolsep}{12pt}
\begin{tabular}{@{}lcc@{}}
\toprule
\textbf{Model} & \textbf{100 images} [$\Lambda \geq 0.2$] & \textbf{1{,}000 images} [$\Lambda \geq 0.1$]\\
\midrule
Asian Biased          & 24 & 83  \\
African/Indian Biased & 80 & 100 \\
Caucasian Biased      & 60 & 100 \\
\midrule
Unbiased              & 10  & 52  \\
Unbiased (+)          & 14 & 2   \\
\bottomrule
\end{tabular}
\end{table}
Both signals concentrate with sample size, as expected for finite-sample statistics: the spread of the Level-2 $KS$ statistic contracts with the evaluation set size and that of the Level-3 activation $\lambda$ sharpens with it. At $100$ images per group, the disparity is already visible---the favored group's average activation is reliably the highest in every biased model---but the \emph{automatic} tests are not yet dependable. The $KS$ $p$-value lacks the power to reject at this size, and the $\Lambda \geq 0.1$ rule flags the strongly biased African/Indian model in $87\%$ of draws, but the weakly biased Asian model in only $24\%$, against $10$--$14\%$ for the two unbiased models---too close to separate reliably. By $1{,}000$ images the detectors are reliable and the Asian model is recovered in $77\%$ of draws, the Afr./Indian and Caucasian models 100\%.
 
At $1{,}000$ images, the activation detector also resolves a subtler effect. The standard unbiased model is now flagged in $52\%$ of draws, whereas the unbiased~(+) model -- trained on three times as much balanced data -- is flagged in only $5\%$. At this sample size, the estimate is tight, so the $52\%$ is not sampling noise: the standard unbiased model carries a small but real representational asymmetry despite its demographically balanced training data, and the additional balanced data of the unbiased~(+) model remove it. Balance in the data therefore does not guarantee balance in the representation -- the representational counterpart of bias encoded in the learned parameters (\S\ref{subsec:res_l4})---and this is the concrete justification for the conservative false-positive budget adopted in the at-scale detection experiments.

\section{Bias Detection Network Ablation}
\label{appx:detector}

\begin{figure}[t]
    \centering
    \includegraphics[width=0.7\columnwidth]{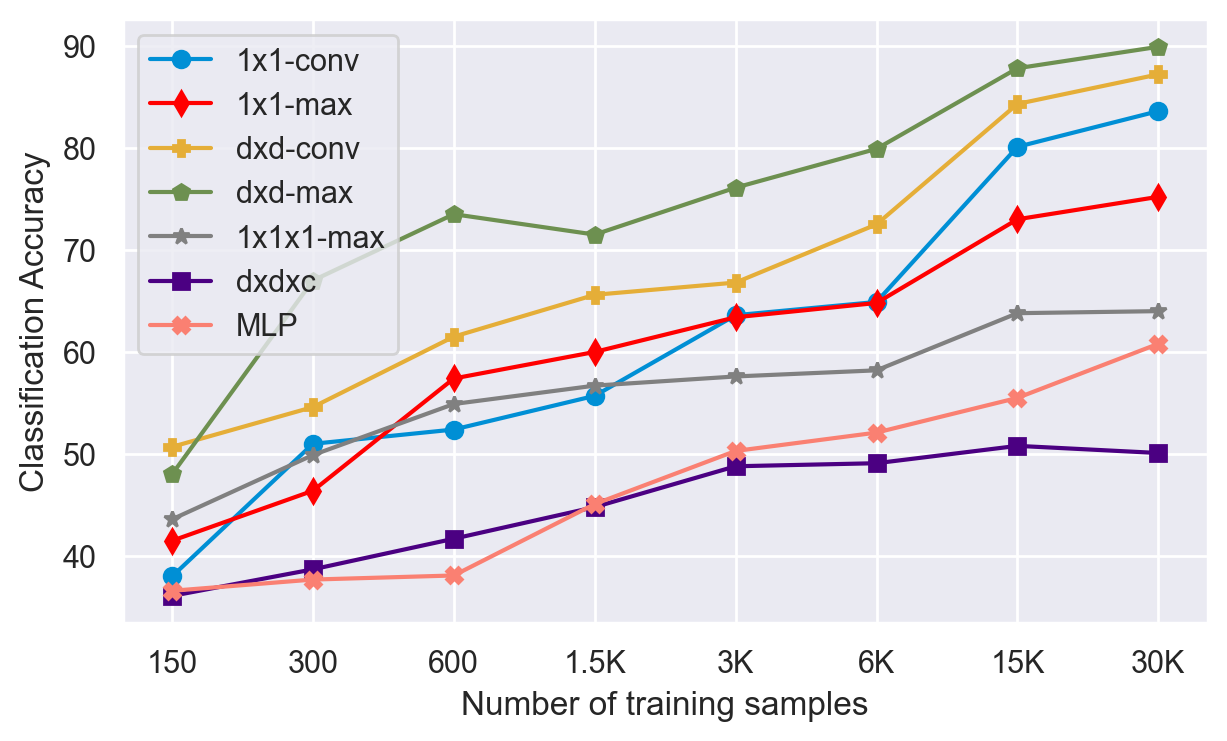}
    \caption{\textbf{Bias classification accuracy} for the different architectures given the number of training samples (x axis). Bias is classified into three different categories according to a demographic criterion, namely Asian, African/Indian, and Caucasian.}
    \label{fig:gender_detect}
\end{figure}

Figure \ref{fig:gender_detect} shows the classification accuracy of various bias detection networks $\psi$ as a function of the number of samples used for training. 

The best performance is obtained with architectures that implement a convolution with the kernel of the size of the input weights ($d\times d$-conv and $d\times d$-max), reaching $87\%$ and $90\%$ classification accuracy, respectively. The improvement in the architecture with the most parameters ($d\times d\times c$, with 5.7 million) stalls at $3$K training samples and above. 

\begin{figure}[t]
    \centering
    \includegraphics[width=0.75\columnwidth]{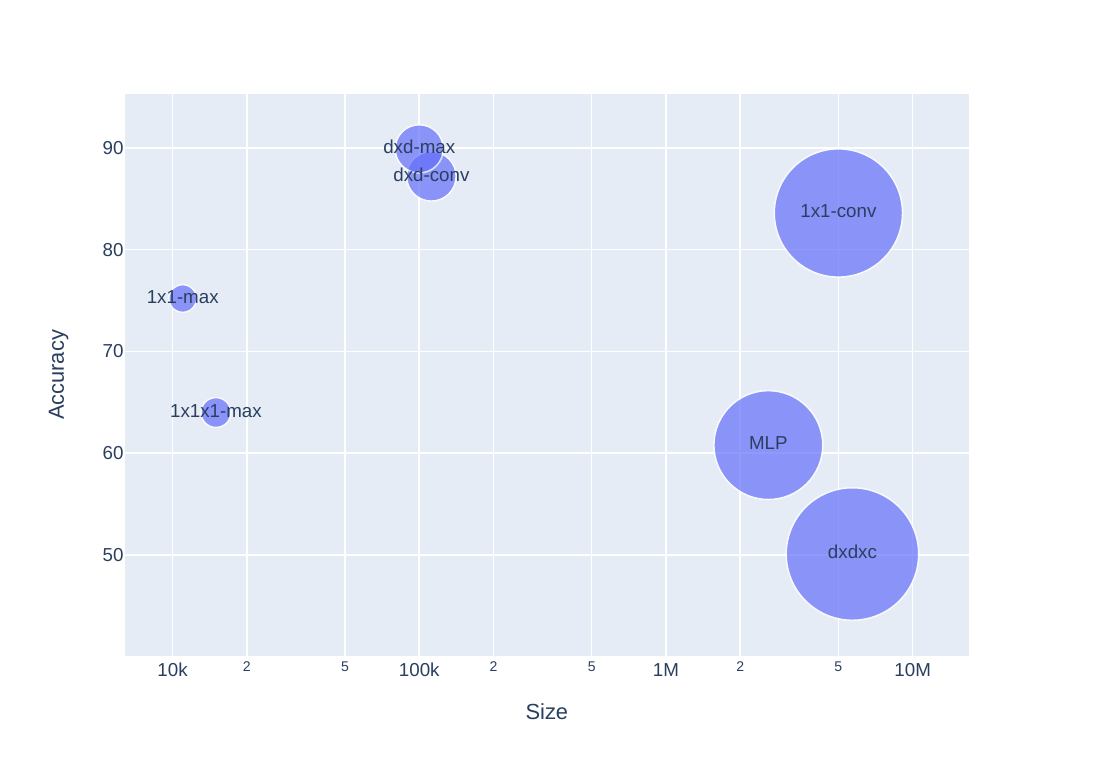}
    \caption[]{\textbf{Bias classification accuracy} for different architectures as a function of the number of  parameters (size). The x axis, in log scale, ranges from 10k parameters to 10M. Each bubble represents a bias detector $\psi$. Bubble size corresponds to model complexity: more parameters, bigger bubbles. Bias is classified into three different categories according to a demographic criterion, namely Asian, African/Indian, and Caucasian. The results demonstrate the existence of identifiable patterns associated with bias in the weights of a trained Neural Network.}
    \label{fig:l4_face_arch}
\end{figure}

Performance scales consistently with the number of training samples across most architectures, including the simple MLP. However, some architectures, such as $1 \times 1 \times 1$-max and $d \times d \times c$, do not appear to benefit significantly from more training samples. Nevertheless, it should be noted that these architectures perform comparably to others when trained with limited data.

The hypothesis that seems to best explain this behavior is that, since there are so many parameters $\mathbf{\Theta}$, the solution space is so large that the choice of a better architectural configuration occurs automatically, leaving unnecessary parameters unchanged as if they were not present \citep{schmidt1992weights}. With little training data, the model $\psi(\cdot | \mathbf{\Theta})$ adjusts very quickly to those data (losses are practically nil) and in just a couple of epochs it no longer needs to adjust those weights $\mathbf{\Theta}$. However, when the number of training samples increases, more parameters are needed to be modified to better correlate the training data, thus losing the generalization capability.

We have dealt with many more architectures that are not worth describing here: using two dense layers at the end, adding a dense layer after each convolution, replacing convolutions with dense layers, playing with dropout, etc.; all resulting in worse performance than the learning architectures reported here.


\end{document}